\documentclass[journal]{IEEEtran}
\usepackage{xcolor}
\usepackage[colorlinks]{hyperref}
\hypersetup{colorlinks,breaklinks,linkcolor=blue,urlcolor=blue,anchorcolor=blue,citecolor=blue}

\usepackage{graphicx}
\usepackage{microtype}
\usepackage{subfigure}
\usepackage{booktabs} % for professional tables
\usepackage{mathtools}
\usepackage{color}
\usepackage{adjustbox}
\usepackage[numbers]{natbib}

\DeclareUnicodeCharacter{00A0}{ }
\usepackage[utf8x]{inputenc} 

\ifCLASSINFOpdf
\else
\fi
\usepackage{algorithm}
\usepackage{algorithmic} 
\usepackage{amsmath}
\usepackage{graphics}
\usepackage{epsfig}

\begin{document}
\title{graph}

\author{ Tianyu Shi$^*$, Jiawei Wang$^*$, Yuankai Wu, Luis Miranda-Moreno, Lijun Sun% <-this %

\thanks{This work is supported by the Natural Sciences and Engineering Research Council (NSERC) of Canada and the Canada Foundation for Innovation (CFI). T. Shi and Y. Wu would like to thank the Institute for Data Valorization (IVADO) for providing scholarships. A preliminary version of this paper was presented at the ICML workshop on AI for Autonomous Driving (AIAD). \textit{Corresponding author: Lijun Sun (lijun.sun@mcgill.ca).}}
\thanks{$^{*}$ Contributed equally.}

\thanks{The authors are with the Department of Civil Engineering, McGill University, Montreal, QC H3A 0C3, Canada.}
}%

\title{Efficient Connected and Automated Driving System with Multi-agent Graph Reinforcement Learning}

\maketitle

\begin{abstract}
Research on connected and automated vehicles (CAVs) has received considerable attention from both industry and academia in recent years. The advances in reinforcement learning have led to substantial progress in controlling a single CAV. However, in mixed autonomy involving both CAVs and human-driving vehicles, it remains a critical challenge to encourage cooperation among CAVs to better regulate human-driving traffic flow and improve the efficiency of the whole transportation system. In this paper, we address the CAV cooperation problem in mixed autonomy by incorporating Connected Automated Vehicle Graph (CAVG) into multi-agent reinforcement learning (MARL) to model the mutual interplay among CAVs. In the proposed framework, CAV cooperation is first learned using graph convolutional networks with attention mechanism. We conduct extensive experiments based on the car-following and unsignalized intersection settings in \textit{flow}. We also evaluate our method on the merging scenarios --- an open road network with a varying number of agents --- to demonstrate the generalization ability. Experimental results show that the proposed MARL-CAVG framework outperforms state-of-the-art baselines for CAV control in improving the efficiency and safety of mixed autonomy transportation system.
\end{abstract}

\begin{IEEEkeywords}
Connected and Automated Driving, Decision making, Multi-agent Reinforcement Learning, Graph Neural Networks.
\end{IEEEkeywords}

% \fi
\IEEEpeerreviewmaketitle

\section{Introduction}
\IEEEPARstart{T}{raffic} system is unstable in nature due to the inherent randomness in human-driving behavior \cite{treiber2013traffic}. Shockwaves and stop-and-go have become a primary safety concern and the main driver for traffic accident. As a promising solution to improve the efficiency of transportation systems, connected and automated vehicles (CAVs) have received increasing attention from both industry and academia. One major benefit of CAVs is that the randomness in driving behavior can be significantly reduced; thus, the whole system can be better managed by control algorithms with minimum reaction time. Theoretically, having a fully autonomous fleet will substantially enhance the capacity and efficiency of urban transportation systems. However, before reaching full autonomy, it is inevitable that both CAVs and human-driving vehicles co-exist and interact with each other. In such a mixed-autonomy system, how to design and optimize CAV behavior becomes critical to the development and implementation of future autonomous driving.

With recent advances in artificial intelligence, reinforcement learning (RL) has become an efficient tool to solve diverse, intelligent control tasks. In particular, RL fits various tasks in autonomous driving, and the field has been greatly advanced thanks to recent development in RL (e.g., flight control \cite{abbeel2007application}, Go games \cite{silver2017mastering}). In particular, RL fits various tasks in autonomous driving, and the field has been greatly advanced thanks to recent development in RL (see, e.g., \cite{wang2018reinforcement}, \cite{wu2017flow}). However, despite these advances, many research challenges remain in promoting CAV cooperation in mixed autonomy. First, as CAVs have different characteristics compared to human-driving agents (e.g., reaction time and action generation process), it becomes challenging to navigate them in such an extremely dynamic and complicated driving environment. Second, in a mixed-autonomy system, it remains unclear how to encourage automated vehicle agents to cooperate to maximize the total expected returns of the whole system. Finally, how to effectively guarantee both safety and efficiency in policy generation is also an urgent research question in a multi-agent automated driving setting.

In this work, we try to address the above challenges by incorporating graph neural networks into a multi-agent reinforcement learning framework to better encourage cooperation. In  doing  so, we integrate information obtained from each CAV into a graph structure---Connected and Automated Vehicle Graph (CAVG), in which the learned latent features are exploited to enhance cooperation among multiple agents. Our main contributions are:
\begin{itemize}
    \item To the best of our knowledge, this work is the first to use the graph attention networks to capture mutual interplay among CAVs in the navigation setting, which is beneficial for cooperative control.
    \item We propose a dynamic adjacency matrix scheme in the decision-making framework to account for both speed and position information, extract valuable information from surrounding neighbors.
    \item We conduct extensive experiments with diverse levels of complexity. We compare the proposed MARL-CAVG model with several state-of-the-art baselines for model evaluation. Our MARL-CAVG model can better balance between safety and efficiency in mixed-autonomy system.
\end{itemize}

The rest of the paper is organized as follows. Section~\ref{sec:review} summarizes the related literature, in particular recent developments in RL. Section~\ref{sec:method} introduces the proposed MARL-CAVG model for CAV control. Extensive experiments based on simulated scenarios are presented in Section~\ref{sec:experiment}. Section~\ref{sec:conclusion} summarizes concluding remarks and  future research directions. 

\section{Related Work} \label{sec:review}
Most existing studies in the field of automated vehicles control and motion planning have been focusing on maximizing the efficiency for an individual agent (i.e., ego driving), which formulates automated motion planning as an optimization problem and then solves it with rule-based models (see, e.g., \cite{rasekhipour2016potential}, \cite{luo2019cooperative}). However, such methods may fail in many real-world scenarios due to the complex interactions among agents. To address the limitation, recent developments for automated driving have been shifted from rule-based methods to reinforcement learning-based models, which are more flexible, efficient, and demonstrate superior generalization power. Meanwhile, integration of micro-traffic simulator such as SUMO \cite{lopez2018microscopic} with deep reinforcement learning library enables easy implementation of different traffic control tasks, e.g., lane change, ramp merge, and intersection \cite{wu2017flow,shi2019driving}. Despite the promising results, existing reinforcement learning-based approaches still mainly focus on the control of a single-agent in a static or fully observed setting. As a result, these methods are still limited to non-shared policy generation rather than exploring multi-agent shared policy and cooperation under mixed autonomy.

Real-world automated driving problems often involve multiple agents in a dynamic and partially observed environment. An emerging question is how to promote cooperation among agents instead of relying on ego/selfish-driving. \citet{shalev2016safe} introduce a hierarchical
temporal abstraction with a gating mechanism that
significantly reduces the variance of
the gradient estimation in multi-agent automated driving environment. Furthermore, \citet{palanisamy2019multi} 
uses Partially Observable Markov Games (POSG) to
formulate the connected automated driving problems which can be trained in both centralized and decentralized framework. 
\citet{wang2019cooperative} develop the cooperative lane change system by considering the overall traffic efficiency instead of the travel efficiency of an individual vehicle, which can lead
to a more harmonic and efficient traffic system rather than
competition.
However, it remains unknown how to better utilize the information of surrounding agents to encourage the cooperation and make the driving behavior more efficient.
 % 我们用shared policy的理由应该是现实车辆数量庞大，使用不share的policy学习困难。但很多自动驾驶车辆都有着共性，所以使用share policy是不错的选择，

Recent research progress on graph information sharing has brought new and promising perspectives to the multi-agent reinforcement learning problems \cite{wu2020comprehensive}. \citet{iqbal2019actor} propose to use multi-head attention mechanism to enable effective and scalable learning in complex multi-agent environments. However, this framework does not consider parameter sharing among neighbors, which may make it hard to train and implement into the CAV setting.  \citet{agarwal2019learning} propose to create a shared agent-entity graph and introduced curriculum learning to increase transferability.
\citet{jiang2018graph} propose the graph convolutional reinforcement learning approach for multi-agent to learn cooperative strategies. However, for a highly dynamic environment, such as the automated driving setting, both vehicles in close-range and vehicles with high relative speed to the ego vehicle should be considered. Previous studies have not fully utilized both position and speed information from surrounding agents, which hinders the feasibility of real-world implementation of CAV.

\section{Methodology} \label{sec:method}
\subsection{CAV Control Framework}
In this paper, we consider the CAV control framework presented in Fig.~\ref{framework}. Following the study of \cite{kreidieh2018dissipating}, we model $N$ CAVs as $N$ homogeneous agents in a mixed-autonomy traffic network to achieve better generalization ability. Their decision procedures can be divided into three stages: (1) in the beginning of each decision, the agent $c_{i},i=1,...,N$ will first have a local observation and identify their current state $s_{i}^t$ at time step $t$; (2) then each agent will manage to locate and sense their neighbors; (3) once the agents acquire both information from themselves and their neighbors, they will decide whether to accelerate/decelerate accordingly.
% \footnote{Because our code is based on Tensorflow, we have to consider a fixed number of agents, if the surrounding neighbor vehicles in our scan scale is less than the agent number, then the feature will be replaced by the 0.} 

\begin{figure}[htbp]
\vskip -0.1in
\begin{center}
\centerline{\includegraphics[width=\columnwidth]{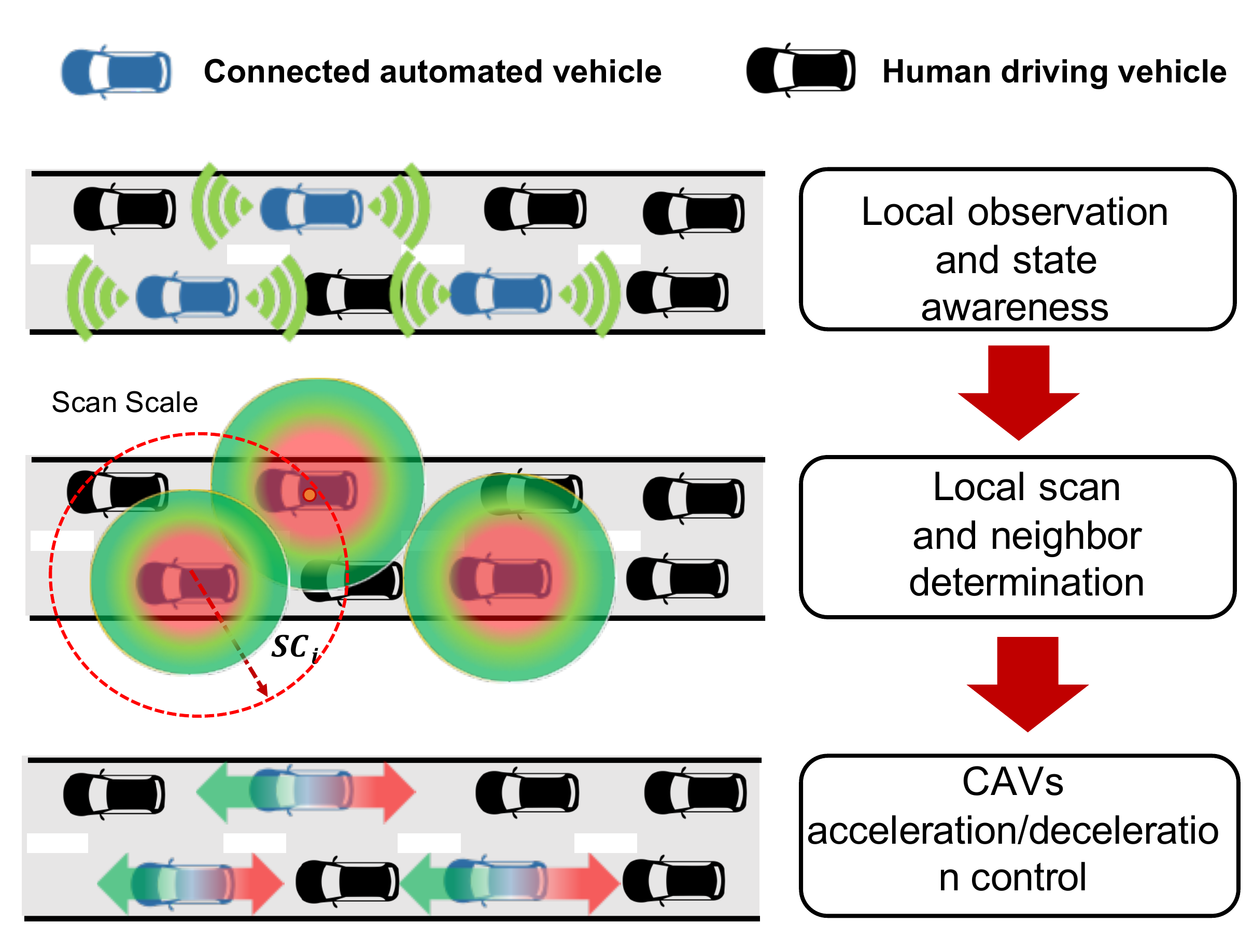}}
\caption{CAV control framework. The blue vehicles represent the CAVs while the black vehicles represent the human-driving vehicles. The colormaps stand for the Gaussian speed field of each CAV. The red dotted line stands for the scan scale which is the local observation range of each agent. } 
\label{controlframework}
\end{center}
\vskip -0.1in
\end{figure}

We first illustrate in detail how the reinforcement learning framework can be utilized in the CAV control problem. Formally, the task in such a mixed-autonomy transportation system can be defined in the setting of multi-agent reinforcement learning (MARL) with the following components:
\begin{itemize}
\item \textbf{Agent}: In the simulation, we mainly consider two types of agents: a) human-driving agents whose acceleration or deceleration decisions are determined based on car-following model (i.e., intelligent driver model \cite{treiber2000congested}); b) CAV agents which are controlled by a deep reinforcement learning framework. 
\item \textbf{State}: We assume each CAV agent $c_i,\  i=1,...,N$ has local observations $o_i$. Specifically, $o_i$ consists of two types of feature: the first is speed and position of $c_i$, which is used to characterize its dynamic in the system; the second is the relative speed and position of $c_i$ with its following CAV agent and leading CAV agents, which can be regarded as local information from other CAV agents. Note that different setting of relative information will be further illustrated and analyzed in experimental studies. Moreover, we consider each agent $c_i$ can communicate with other CAV agents, therefore the overall state observation of the $i$-th CAV is denoted by $S_i = \{o^{1}_i,..., o^{m}_i\}, i=1,..., N$, where $m$ is the maximum number of neighboring CAV agents which $c_i$ can sense. In this way each CAV agent can develop more cooperative policy by incorporating information from neighbors before making control decision.

% \footnote{Note that for all scenarios, we run 100 episodes with 10 random seeds. And for some scenarios, we found that the information from follower vehicle will make little difference on overall performance, so we did not include these.}

\item \textbf{Action}: Action $a_i,\  i=1,..., N$ is the speed adjustment of CAV agents. It is bounded by the maximum allowable acceleration $a_{acc}$ and allowable deceleration $a_{dec}$ (i.e., $a_i \in [a_{dec},a_{acc}], a_{dec}<0, a_{acc}>0$).

\item \textbf{Reward function}: We define specific reward functions for different simulation scenarios. 

For the \textit{ring} scenario and the \textit{figure eight} scenarios, we define the reward function to encourage high average speeds from all vehicles in the network and penalize accelerations/decelerations by the CAVs. 
The rewards for \textit{ring} and \textit{eight} scenario are defined as:
\begin{equation}
r_i = -w_v \times (\Bar{v}_{T}-\Bar{v}) + w_a \times(\hat{a}-\Bar{a}),
\label{rewardringandeight}
\end{equation}
where $w_v$, $w_a$ stand for the weight parameters for average velocity and average acceleration which are predefined as 2 and 4 respectively. Note that  $\Bar{v}_{T}$ is the target velocity and  $\hat{a}$ is the threshold of the acceleration.

For the \textit{merge} scenario, reward function encourages the consistency between the system-level speed and desired speed, while slightly penalizing short headways among CAVs \cite{wu2017flow}:
\begin{equation}
r^t_i = -w_v \times (\Bar{v}_{T}-\Bar{v})+ w_h \times( \text{min}(\frac{(\Bar{h}-t_{min})}{t_{min}},0)),
\label{rewardmerge}
\end{equation}
where $w_v$ and $w_h$ denotes the weight parameters for average velocity $\Bar{v}$ and average headway $\Bar{h}$ which are predefined as 1 and 0.1 respectively. In addition, there is a smallest acceptable time headway $t_{min}$, which is defined as 1 in this study.

\item \textbf{Termination}: An episode is terminated if the time horizon is reached or a collision happens.

\end{itemize}

\subsection{Multi-agent Cooperation Within Mixed-autonomy System} 

In most previous literature (e.g., in \cite{shi2019driving}, \cite{wang2018reinforcement}), each CAV maintains its own policy in the environment. This is not efficient for controlling a group of CAVs in mixed-autonomy transportation environment due to training complexity. 
Therefore, we introduce the shared policy into our control framework. In this way, we can not only exploit experience from different CAVs as they encounter different situation, but also develop a common sense to efficiently navigate these vehicles in the mixed-autonomy system.  

\begin{figure}[ht]
\vskip 0.2in
\begin{center}
\centerline{\includegraphics[width=\columnwidth]{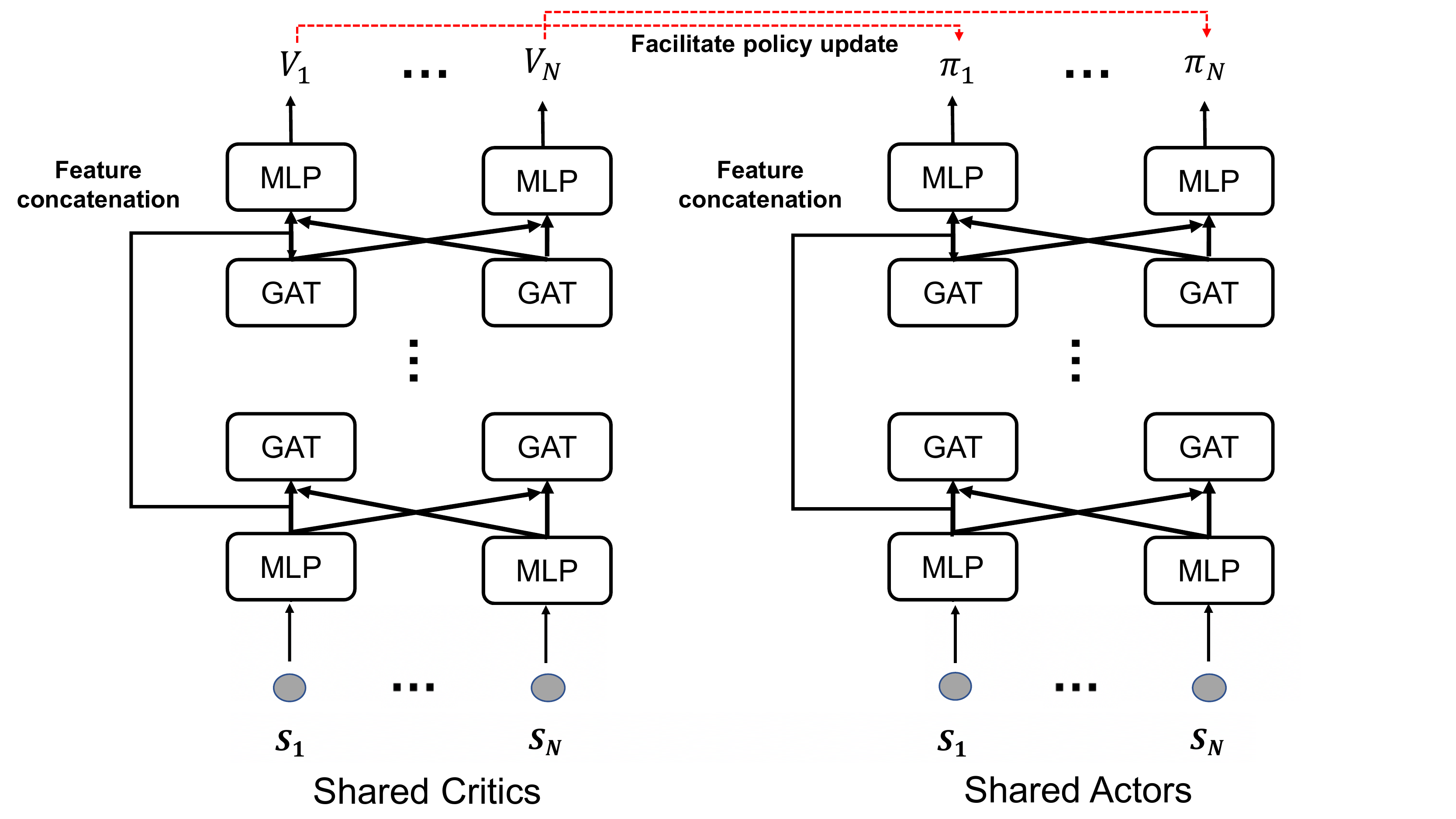}}
\caption{The architecture of MARL-CAVG. The critic network is a graph convolutional neural network parametrized with $\phi$. Notably, the output of critic network at each time step $t$ is state value estimation $V_i$ (i.e., short for $V(S_t,M_i)$), it will be further used for advantage estimation to train the actor network.}
\label{dgnppo}
\end{center}
\vskip -0.1in
\end{figure}

% MARL+CAVG
To better capture neighboring information, we establish a graph attention based multi-agent reinforcement learning architecture as shown in Fig.~\ref{dgnppo}. Specifically, CAVs learn their policies with PPO as the basic optimization scheme to handle continuous action space. The overall architecture is based on the actor-critic algorithm \cite{sutton2018reinforcement} as shown in Fig.~\ref{dgnppo}.

In this study, we design and integrate several techniques to encourage cooperation to promote safety and efficiency in dynamic traffic flow:
\subsubsection{Capture Mutual Interplay among CAVs}
In connected and automated driving scenarios, the environment is extremely dynamic because agents keep moving and their relationships among neighbors change quickly. This characteristic makes it very difficult for agents to learn to cooperate with each other.

\begin{figure}[H]
\vskip -0.1in
\begin{center}
\centerline{\includegraphics[width=\columnwidth]{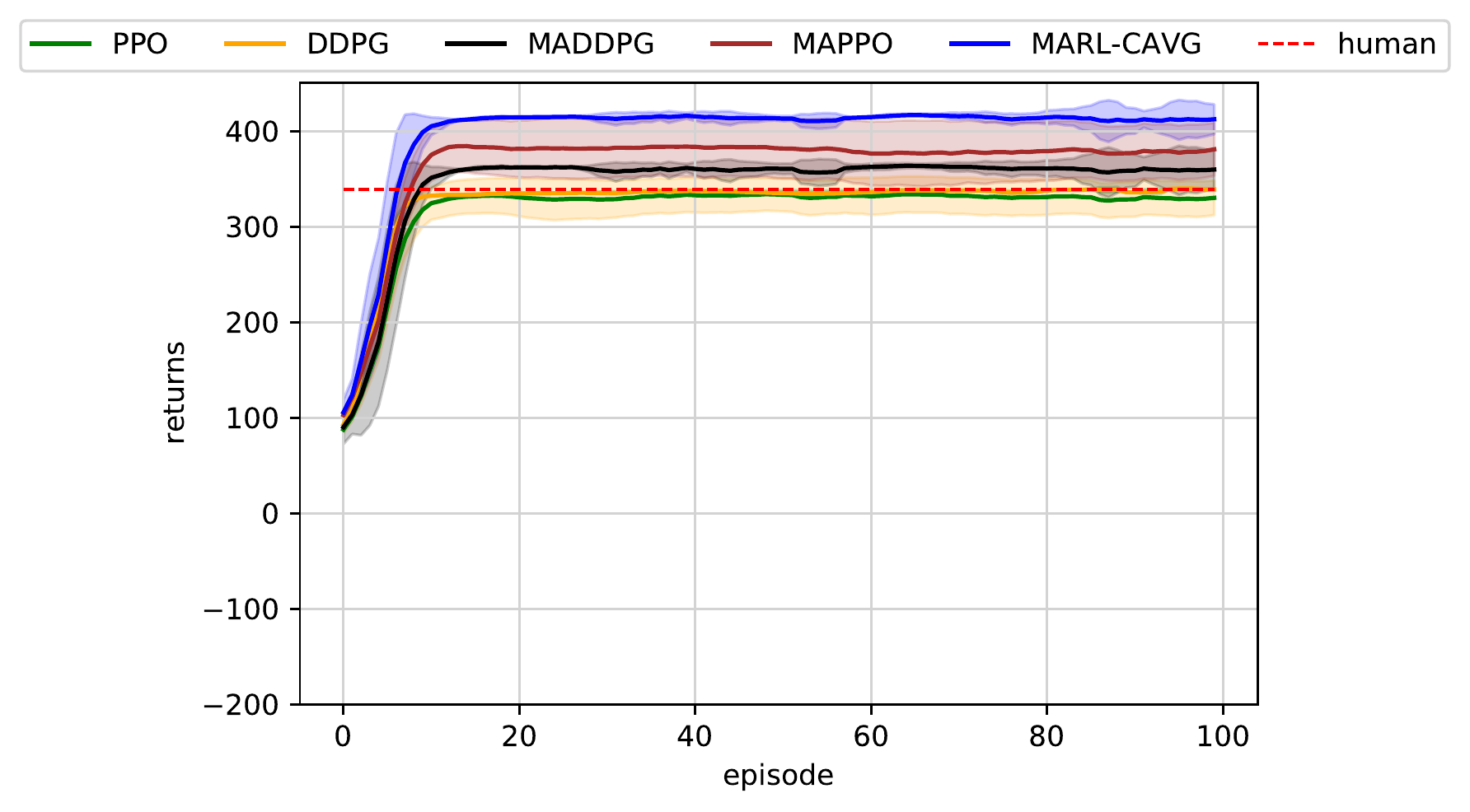}}
\caption{Learning curves in merge network. We use 600 horizon, we consider the penetration rate as 0.25, i.e., 25\% of the vehicles are autonomous vehicles, the number of lane is 1, the target speed is 30km/h. } 
\label{framework}
\end{center}
\vskip -0.1in
\end{figure}

As shown in Fig.~\ref{framework}, after observing the local state, the CAVs will integrate information from their neighbors to develop a more comprehensive awareness of the current traffic dynamic. Firstly, unlike previous approaches (e.g., \cite{jiang2018graph}, \cite{wei2019colight}), we build the adjacency matrix based on Gaussian speed field using Gaussian process regression (GPR) model \cite{zhang2020spatiotemporal}. Compared to only using relative position or velocity information, the advantage of using GPR is to capture the spatial and temporal interactions among surround agents and  adaptable to the dynamic changing environment. The standard exponential kernel function is computed as:
\begin{equation}
K(x_i,x_j) = A \cdot \exp\left(-\frac{(x_i-x_j)^2}{2\sigma^2}\right),
\label{speed}
\end{equation}
where $A$ is an amplitude constant. $x_i$ represents the position of the ego vehicle and $x_j$ represents the position of the surrounding $j^{th}$ vehicle. And $\sigma$ is length scale constant controlling how the correlations are decaying with respect to the distance. A small $\sigma$ indicates fast decay rate, which imposes less correlation on two points that are far away. In our research, we fixed the length scale as 4 meters as suggested by \cite{zhang2020spatiotemporal}.

Furthermore, we can dynamically construct the feature representation in the adjacency matrix at time step $t$, the posterior distribution of the relative velocity $\Delta V(x_i,x_j)$ between the $i$-th CAV agent and its surrounding vehicle $j$ at position $x_i$ can be used to model the interaction relationship among vehicles~\cite{zhang2020spatiotemporal}. These elements are defined as follows:

\begin{small}
\begin{equation}
M_t(i,j) = K(x_i,x_j)K(x_i,x_j)^{-1}\Delta V(x_i,x_j),dist(x_i,x_j) \leq SC_i,
\label{feature}
\end{equation}
\end{small}

We incorporate the location information (i.e., $K(x_i,x_j)$) and velocity difference (i.e., $\Delta V(x_i,x_j)$) of every two agents in the suggested adjacency matrix.
% {\color{red} Notably, each row $i$ represents $i$-th CAV's feature, and each non-zero element of this row is the neighboring information between itself and surrounding vehicles within its scan scale $SC_i$. }
Intuitively, the $i$-th CAV agent will be more sensitive to a closer surrounding vehicle with higher relative speed than a more distant one with lower relative speed (see Fig.~\ref{controlframework}).

Intuitively, the observation and extracted features of each agent are integrated through graph convolution based on the weighted adjacency matrix $M_t$:
\begin{equation}
h^{k}_i = f\left(\text{concat} \left[M_{t}H^{k-1}, D^{-1}_{i}M_{t}H^{k-1}\right]W_{i}\right),
\label{eq:gc}
\end{equation}
where $f$ is the activation function, $D$ is the degree matrix and $h^{k}_i$ denotes extracted feature by agent $i$ at the $k$ th layer, which depends on the current adjacency matrix $M_t$ as well as the feature of its neighbors extracted from previous layer $H^{k-1}=\left[h^{k-1}_1,...,h^{k-1}_N\right]$. 

%% attention
Furthermore, an attention module is added to capture the impact of the surrounding agents. The neighbors are selected within the scan scale $SC_i$ for each CAV individually. Considering  $\mathcal{N}_i$ neighbors of ego CAV $i$, the attention based feature selection on neighboring CAV $j$ can be computed as:
\begin{align}
    q_i&=f^{\text{query}}\left(h_i*W^{\text{query}}\right), \label{eq:query1} \\
    k_j&=f^{\text{key}}\left(h_j*W^{\text{key}}\right),\ \  j\in\mathcal{N}_i \label{eq:query}\\
    v_j&=f^{\text{value}}\left(h_j*W^{\text{value}}\right),\ \  j\in\mathcal{N}_i \label{eq:value} \\
    \phi_{ij} &= \text{softmax}\left(\frac{q_i*k_j^T}{\sum_{l\in \mathcal{N}_i}q_i*k_l^T}\right),
\label{eq:att_score} \\
h'_i &= \sum_{j\in\mathcal{N}_i}\phi_{ij}*v_j.
\label{eq:feat} 
\end{align}

Note that we omit the layer index here for simplicity. $\mathcal{N}_i$ denotes the set of agent $i$ as well as its selected neighbors. We use $f^{\text{query}}$ and $f^{\text{key}}$ to encode input features as query-key pairs, then we implement softmax on dot-product between query and key vectors, with which we can quantify the attention score $\phi_{ij}$ of  agent $i$ on neighboring agent $j$ \cite{vaswani2017attention}. Based on the value $v_j$ and  $\phi_{ij}$, new feature vector $h'_i$ for agent $i$ based on its neighbors can be derived by Eq.~\ref{eq:feat}. With this attention scheme, an CAV agent can further utilize information from neighboring CAVs selectively, and thus the framework can promote more effective cooperation.

\subsubsection{Continuous Action Generation via Proximal Policy Optimization}
% MARL+CAVG
In a typical reinforcement learning problem, an agent takes action $a\in A$ based on the current state $S$ and acquires the reward $R$. Unlike previous tasks based on DQN ( e.g., in Go games \cite{silver2017mastering}), the CAVs need to generate continuous action space for smooth and efficient control strategy.
Therefore, we use Proximal Policy Optimization (PPO) \cite{schulman2017proximal} for CAVs to handle continuous action space. We design the critic network as a graph convolutional neural network parameterized by $\phi$. Notably, the output of critic network at each time step $t$ is state value estimation $V_i$ (i.e., short for $V(S_t,M_i)$), it will be further used for advantage estimation to train the actor network. 

The update of gradient for critic is based on Temporal-Difference (TD) learning \cite{sutton2018reinforcement}, which can be formulated as:
\begin{equation}
\begin{split}
\nabla_ {\phi}L(\phi) = \nabla_{\phi}E\bigg[\sum_{n=1}^{N}(& r^{t}_i+\hat{V}(S_{t+1},M_i) - \\
& \hat{V}(S_{t},M_i))^2  \bigg].
\end{split}
\label{td}
\end{equation}
The policy $\pi_i$ (i.e., short for $\pi(a_i|S,M_i)$) can be modelled as a distribution (i.e., Gaussian distribution for continuous control) and also parameterized through the graph convolutional network with parameters $\theta$. Therefore at given time step $t$, the policy gradient can be derived with the advantage $\hat{A}^t_i,i=1,..,N$ from critic:
\begin{equation}
\label{eq:pg}
\begin{split}
\nabla_{\theta} J(\theta) =\nabla_{\theta} E_{\pi_{\theta^{old}}} \Bigg[\sum_{i=1}^{N} & \text{min}\bigg( r^t_i(\theta)\hat{A}^t_i,\\
&\text{clip}\big(r^t_i(\theta), 1-\epsilon, 1+\epsilon  \big) \hat{A}^t_i \bigg) \Bigg],
\end{split}
\end{equation}
where the likelihood ratio $r_i(\theta)= \frac{\pi_{\theta_i}(a_i|S_t,M_t)}{\pi_{\theta^{old}}(a_i|S_t,M_t)}$ , and this is done by defining the policy loss function to be the minimum between the standard surrogate loss and an $\epsilon$ clipped parameter. It should be pointed out that, for model simplicity, the adjacency matrices are kept the same for the next state value prediction. This assumption makes sense since the variation is limited between two consecutive state observations, especially when the experiment is studied in fine granularity (i.e., simulation resolution is less than 1 s). In addition, we perform on-policy roll-out to collect the experience (i.e., $\{o_i, a_i,r_i\}$) and the advantage estimation for agent $i$ at time step $T$ is calculated as:  $\hat{A}^T_i = \sum_{t}^{T}\gamma^{t}r^{t}_i - \hat{V}(S_t,M_t)$. 
The overall training algorithm is summarized in Algorithm \ref{alg:A1}. Note that we adopt the centralized training and decentralized execution framework. We assume that each automated vehicle will have a centralized platform (e.g., cloud platform) which can share information among automated vehicles. And each automated vehicle will have its own control strategy based on local observation during execution.
\begin{algorithm}[!ht]
\caption{Training algorithm for CAVs control based on traffic simulation}
 \label{alg:A1}
\begin{algorithmic}
 \STATE {Set time horizon $T$ steps for each simulation, set scan scale $SC$ for all the agents.}
\STATE {Initialize memory buffer $\mathcal{B}=\emptyset$, batch size as $b$. }
 \STATE {Initialize parameters $\phi$, $\theta$ for critic and actor networks.}

 \FOR {each episode}

 \FOR {$t =1$ to $T$}

 \STATE Obtain local state observation $o_i,i=1,...,N$ and global observation $S=\{o_1,..,o_N\}$.

 \FOR {CAV $i = 1,...,N$}
\STATE Obtain adjacency matrix $M_i$ based on given scan scale $SC$.
\STATE Sample action $a_i$ from $\pi^{old}(S,M_i|\theta)$  to control CAV $i$.
 \ENDFOR

 \STATE Obtain next local state observation $o'_i, i=1,...,N$ and global observation $S'=\{o'_1,..,o'_N\}$ as well as the reward signal $r_i$ for each CAV .

 \STATE $\mathcal{B} \leftarrow \mathcal{B} \cup \ \left(a_i,o_i,o'_i,r_i, M_i\right)_{i=1}^N$ 

 \IF {$|\mathcal{B}| \% b = 0$}
 \STATE Fetch experience from $M$ and perform roll-out.
 \STATE Update $\phi$ based on Eq.~\ref{td}
 \STATE Update $\theta$ based on Eq.~\ref{eq:pg}
  
  $\mathcal{B}= \emptyset $
 
 \ENDIF

 \IF{collision happens}
\STATE Break
 \ENDIF

\ENDFOR

\ENDFOR
\end{algorithmic}
\end{algorithm}

\section{Experiments} \label{sec:experiment}
\subsection{Experiments Setup}
\begin{figure*}[htbp]
\centering
\includegraphics[scale=0.6]{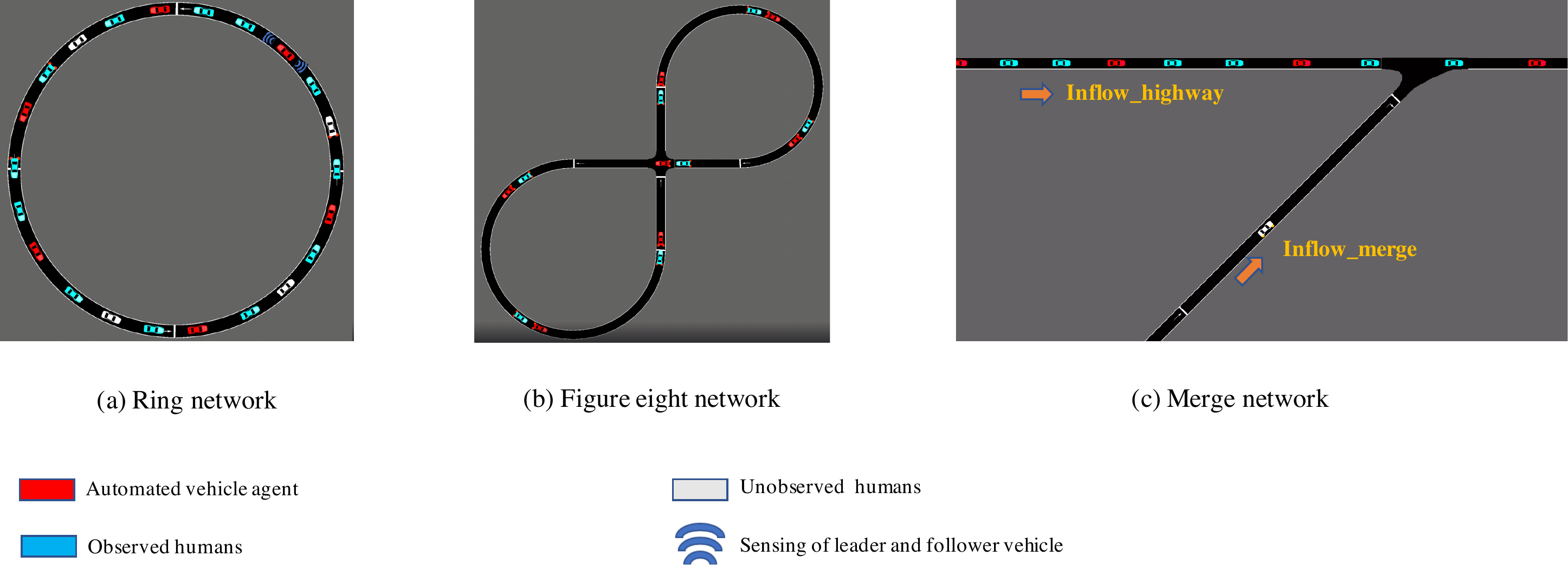}
\caption{Road network structures. To be specific, we consider several variants of simulation scenarios to test the performance of our model.  Firstly, for ring network, it's a standard car-following evaluation scenario which is common in the real world. Secondly, for the figure eight network, it's a more challenging scenario compared to the ring network since it has an intersection. Thirdly, for merge network, which is an open-looped network. As a result, it can be used to test the robustness of our method in a dynamic changing environment. %太长了，只需要形容红色车，绿色车和白色车是什么就可以
}
\label{road-network}
\end{figure*}

We conduct experiments in  Flow\footnote{\url{https://github.com/flow-project}}, an open-source project that supports mixed-autonomy control. Our algorithm is evaluated based on benchmark \textit{car-following} \cite{kreidieh2018dissipating}, \textit{intersection} and \textit{merge} \cite{vinitsky2018benchmarks} as shown in Fig.~\ref{road-network}, which are common intelligent traffic control scenarios. 
In the simulation, horizon represents the number of steps per roll-outs, and each time step is 0.1s.

\begin{itemize}
    
    \item \textbf{Car-following control}. This is a common scenario in the highway without bottleneck.  To simplify the training process, we consider a ring-shape network with single lane which is shown in Fig.~\ref{road-network}-(a). 
    In the initial condition, all the vehicles are uniformly distributed on the circular road with the 
    same initial speed. Experimental results in \cite{sugiyama2008traffic} show that the system is very unstable, even a tiny fluctuation can grow, and eventually breaks up the homogeneous movement, resulting in a traffic jam.
    \item \textbf{Intersection control}. This is a common urban traffic scenario. In this case, CAV control can help improve the overall efficiency of urban transportation systems. We consider a simple intersection with a figure-eight shape network with one or two circular tracks. (see Fig.~\ref{road-network}-(b)).
% https://github.com/flow-project/flow/blob/a511c41c48e6b928bb2060de8ad1ef3c3e3d9554/flow/envs/multiagent/ring/accel.py#L155
    \item \textbf{Merge}. This scenario is common in highway networks. Vehicles move from the on-ramp create backwards propagating stop-and-go waves. As a result, perturbations will propagate upstream from the merge point and reduce the throughput of vehicles in the network. For the merge scenario, total number of vehicles is considered as the number of vehicles per hour coming into the highway lane. See Fig.~\ref{road-network}-(c) for the implemented merge scenario.
\end{itemize}

% \begin{table}[!h]
% 	\centering
%     \setlength\tabcolsep{2pt}% 调整列间距
%     %\\addtolength{\\tabcolsep}{-5pt}% 调整列间距
% 	\footnotesize
% 	\caption{Main Parameters of Different Scenarios}
% 	\scalebox{0.92}{\begin{tabular}{ccccccc}\hline\hline \noalign{\smallskip}
% 		Scenarios & Horizon  & Total vehicles &Penetration rate &Lanes &Target speed(km/h)\\ %
% 		\noalign{\smallskip}\hline\noalign{\smallskip}
% 		Ring-v0 & 3000  &22  &0.18&1&30\\
% 		\noalign{\smallskip}\hline\noalign{\smallskip}
% 		Ring-v1 & 3000 &22  &0.55&1&30\\
% 		\noalign{\smallskip}\hline\noalign{\smallskip}
% 		Ring-v2 & 3000  &22  &0.77&1&30\\
% 		\noalign{\smallskip}\hline\noalign{\smallskip}
% 		Eight-v0   & 1500  &14 &0.50&1&25\\
% 		\noalign{\smallskip}\hline\noalign{\smallskip}
% 		Eight-v1   & 1500  &14 &0.50&2&25\\
% 		\noalign{\smallskip}\hline\noalign{\smallskip}
% 		Eight-v2   & 1500  &14 &0.50&1&40\\
% 		\noalign{\smallskip}\hline\noalign{\smallskip}
% 		Merge-v0 & 600    & 1000  &0.25&1&30\\
% 		\noalign{\smallskip}\hline\noalign{\smallskip}
% 		Merge-v1 & 600    & 1500  &0.25&1&30\\
% 		\noalign{\smallskip}\hline\noalign{\smallskip}
% 		Merge-v2 & 600    & 2000  &0.25&1&30\\
% 		\noalign{\smallskip}\hline\hline
% 	\end{tabular}}
% 	\label{param}
% \end{table}

\subsection{Algorithms Setup}
We compare our MARL-CAVG method with several state-of-the-art baselines, including not only reinforcement learning frameworks (single-agent and multi-agent) but also car-following models from traffic flow theory. For all experiments, we run 100 episodes with a collection of the average results of 10 random seeds. The explanations for selecting these baselines are given as follows. 

\textbf{Intelligent driver model (IDM)}:
\begin{itemize}
    \item Intelligent driver model (IDM) \cite{treiber2000congested} is a commonly used adaptive cruise control method for vehicles that automatically adjusts the acceleration based on distance and velocity information to maintain a safe distance from the leading vehicle. IDM is commonly used to model human-driving behavior in traffic simulators. (e.g., \cite{wang2018reinforcement,shi2019driving}). We consider to add 0.2 random noise on the action to model the uncertainty of human-driving behavior.
\end{itemize}   
\textbf{Single-agent methods}:
\begin{itemize}
    % Need to update the statement for ddpg and mappo
    % is ddpg multiagent?NO
    % mappo is not based on maddpg framework.
    \item Deep Deterministic Policy Gradient (DDPG) \cite{lillicrap2015continuous}: DDPG is a deterministic version of model-free  RL algorithm to deal with continuous action space. CAV agent can reliably learn the optimal policy with continuous actions. We construct a single-agent training framework based on DDPG method, which is similar to \cite{huang2019autonomous}.
    \item Proximal Policy Optimization (PPO) \cite{schulman2017proximal}: PPO is a gradient-based RL algorithm to deal with continuous action space. Unlike DDPG, PPO is an on-policy algorithm.
\end{itemize}
\textbf{Multi-agent methods}:
\begin{itemize}
    \item Multi-agent Deep Deterministic Policy Gradient (MADDPG)~\cite{lowe2017multi}: This is a widely used multi-agent framework with centralized critic and decentralized actor. This is a baseline model which does not specifically consider information from neighbors with graph attention mechanism.
    \item Multi-agent Proximal Policy Optimization (MAPPO): This framework is developed based on the single-agent version of PPO. Unlike in single-agent PPO, different agents will have a shared policy in MAPPO.

\end{itemize}

\subsection{Performance Comparison}

We conduct several experiments in these three networks, the simulation recordings can be found here~\footnote{\url{Simulation : https://youtu.be/86LNKyyqZPQ.}}.

\begin{figure}[H]
\vskip -0.1in
\begin{center}
\centerline{\includegraphics[width=1\columnwidth]{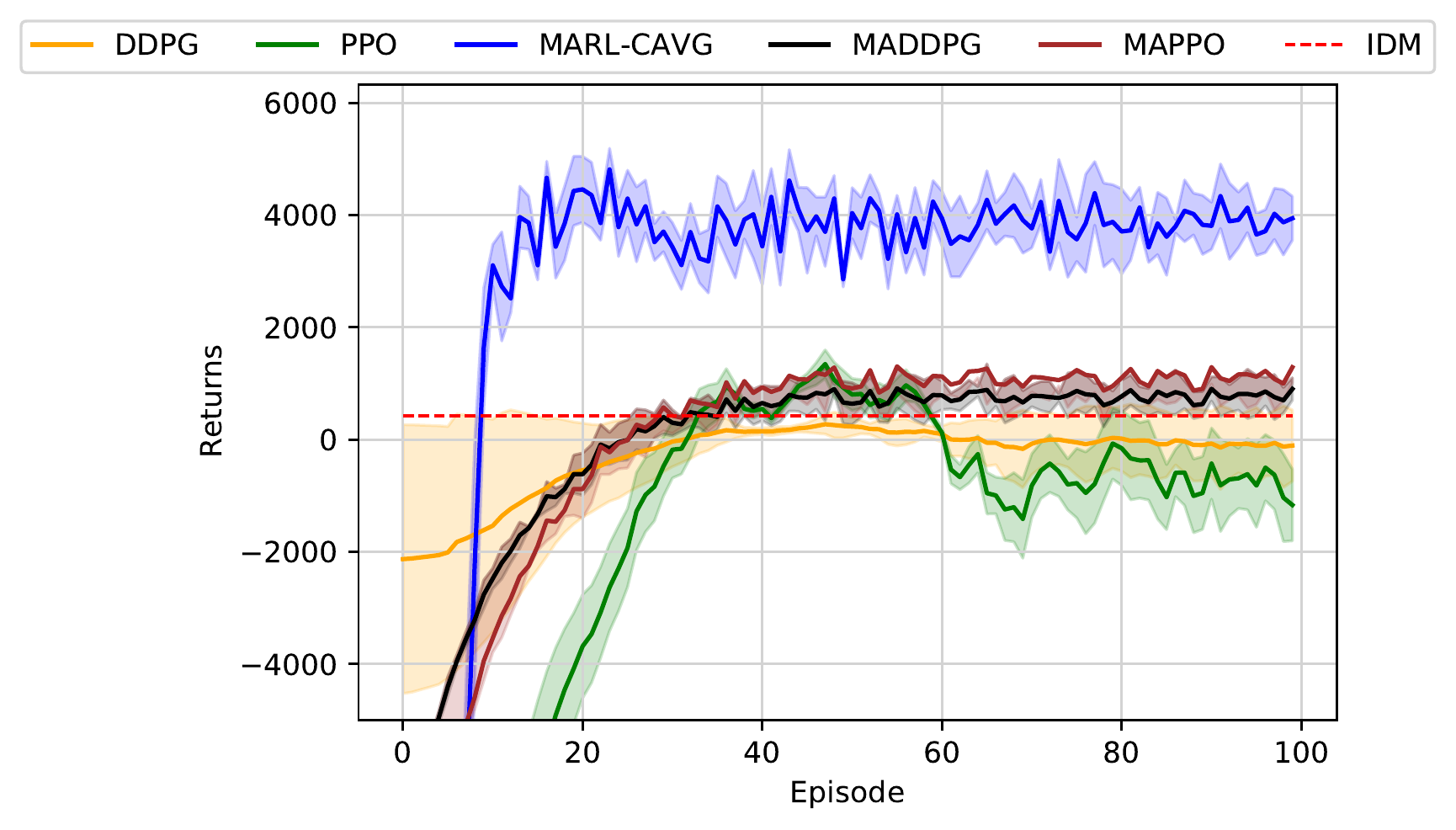}}
\caption{Learning curves in ring network. In ring network, we use 3000 horizon, the total vehicles are 22 and autonomous vehicles are 16. The lane is 1 and target speed is 30km/h.} 
\label{ringreturn}
\end{center}
\vskip -0.1in
\end{figure}

\noindent\textbf{Evaluation in Car-Following Control.}  Fig.~\ref{ringreturn} shows the training performance of different methods in the car-following control scenario. As we can see, the MARL-CAVG method outperforms other methods with a large margin. We can see from Table~\ref{ringtable} that MARL-CAVG achieves the second highest velocity and the smallest acceleration, which makes it achieve highest return in this scenario. It indicates that our model can better mitigate the shock-wave during the car following control.

% analyze penetration rates
%  We change penetration rate as 0.18, 0.55, 0.77 in \textit{ring-v0},\textit{ring-v1},\textit{ring-v2}. From the simulation recordings\footnotemark[2] and obtained results in Fig.~\ref{results}, we have several interesting findings. Firstly, for the \textit{ring} scenario, we find that with small penetration rate in \textit{ring-v0}, MARL-CAVG method has similar results as multi-agent methods. However, with the increase of penetration rates, both multi-agent methods and single-agent methods become worse than the MARL-CAVG method. The declination of multi-agent methods is even larger than single-agent methods. The reason is that,  for  multi-agent methods, the uncertainty will increase with the number of agents because of policy sharing. Thanks to the graph attention mechanism from surrounding neighbors, MARL-CAVG method still demonstrates superior performance with increasing penetration rates.

\begin{table}[htbp]
	\centering
	\footnotesize
	\caption{Performance comparison in ring network}
	\begin{adjustbox}{max width=0.45\textwidth}

	\begin{tabular}{cccccc}\hline\hline \noalign{\smallskip}
		Methods & Velocity ($m/s$) &Acc (0.1 $m^2/s$) & Return\\ %
		\noalign{\smallskip}\hline\noalign{\smallskip}
		IDM   & 2.754   & 3.318 &424.12\\
		\noalign{\smallskip}\hline\noalign{\smallskip}
		DDPG   & 3.134 ($\pm$0.148)  & 2.718($\pm$0.378) &-70.063($\pm$70.506)  \\
		\noalign{\smallskip}\hline\noalign{\smallskip}
		PPO   & 3.165 ($\pm$0.145)  & 3.129 ($\pm$0.369)&-676.959($\pm$65.046) \\
		\noalign{\smallskip}\hline\noalign{\smallskip}
		MADDPG   & 3.270 ($\pm$0.148)   &2.121 ($\pm$0.366) &779.140($\pm$29.178)  \\
		\noalign{\smallskip}\hline\noalign{\smallskip}
		MAPPO   &  3.379 ($\pm$0.142) & 2.782 ($\pm$0.325) &776.225($\pm$20.121) \\
		\noalign{\smallskip}\hline\noalign{\smallskip}
		MARL-CAVG   & \textbf{3.391} ($\pm$0.092)  & \textbf{0.835 }($\pm$0.171) &\textbf{2593.99}($\pm$51.820)  \\
		\noalign{\smallskip}\hline\hline
	\end{tabular}
	\end{adjustbox}
	\label{ringtable}
\end{table}

%% add comparison metric

\begin{figure}[H]
\vskip -0.1in
\begin{center}
\centerline{\includegraphics[width=0.9\columnwidth]{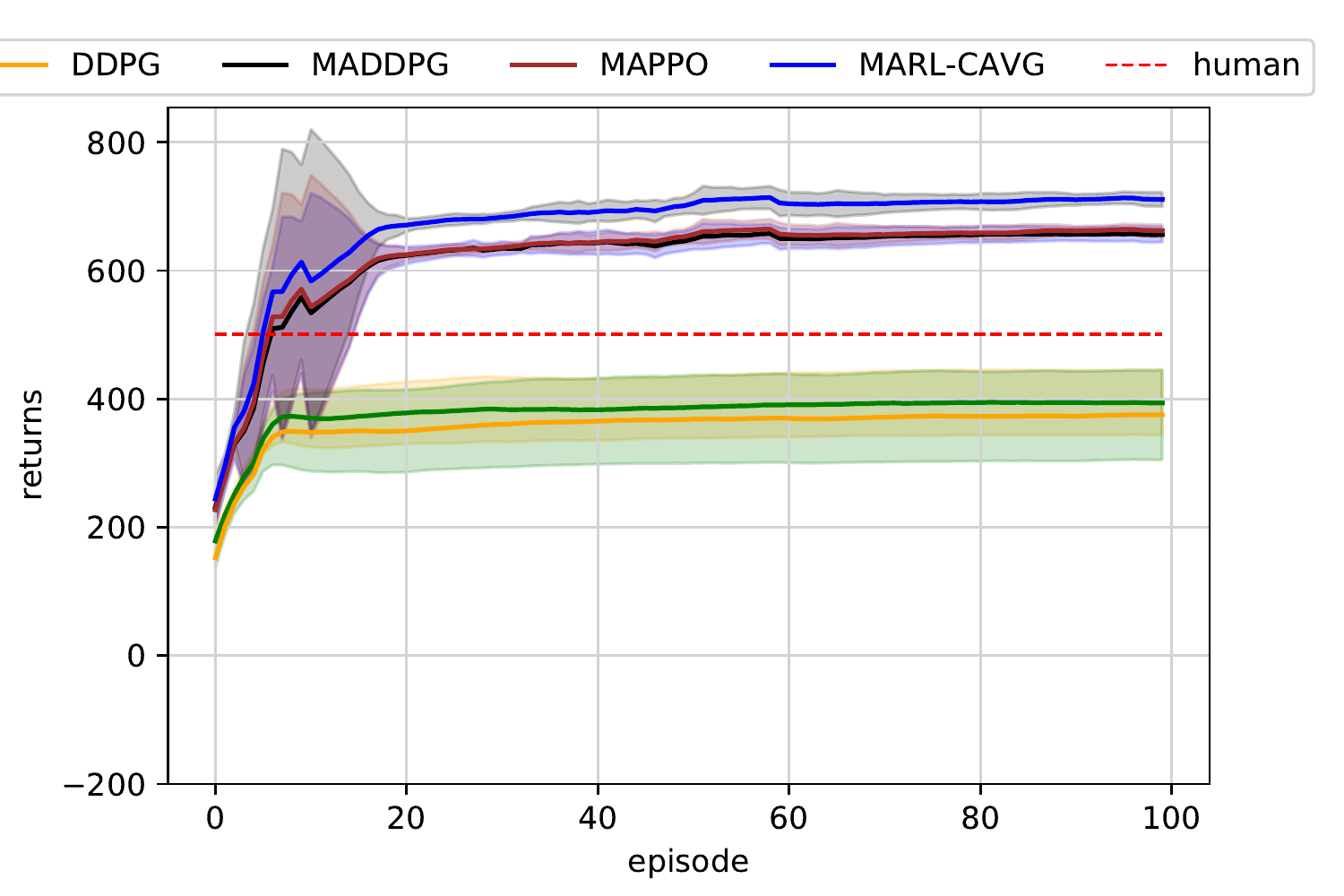}}
\caption{Learning curves in figure eight network. In figure eight network, we use 1500 horizon, the total vehicle number is 14, and the autonomous vehicle number is 7, and the target speed is 30km/h.} 
\label{eightreturn}
\end{center}
\vskip -0.1in
\end{figure}

\noindent \textbf{Evaluation in Intersection Control.} Secondly, in the \textit{eight} scenario, the objective is to balance the safety and efficiency in the figure eight network. To be specific, with the introduction of lane change behaviors or increasing target speed, the average speed within the network will increase and therefore the possibility for collisions at the intersection will be higher. As shown in Fig.~\ref{eightreturn}, we find that MARL-CAVG method can achieve better performance when considers the aforementioned changes. It can achieve better control performance, maintaining a good balance between safety and efficiency. From Table~\ref{eighttable}, we can find that although MARL-CAVG does not achieve the highest velocity, it has the smallest acceleration in this scenario. This demonstrates that our model can learn to sacrifice the speed but achieve higher safety to achieve the optimal trade-off, which is beneficial to get the highest cumulative return.
 
\begin{table}[H]
	\centering
	\footnotesize
	\caption{Performance comparison in figure eight network}
	\begin{adjustbox}{max width=0.45\textwidth}
	\begin{tabular}{cccccc}\hline\hline \noalign{\smallskip}
		Methods & Velocity ($m/s$) &Acc (0.1 $m^2/s$) & Return\\ %
		\noalign{\smallskip}\hline\noalign{\smallskip}
		IDM   & 4.531   & 9.205  &500.87 \\
		\noalign{\smallskip}\hline\noalign{\smallskip}
		DDPG   &   4.325($\pm$2.091) & 8.619($\pm$3.191)  &379.061(($\pm$46.852) \\
		\noalign{\smallskip}\hline\noalign{\smallskip}
		PPO   &  4.064($\pm$2.415) &   8.812($\pm$4.090) &357.946($\pm$64.509)\\
		\noalign{\smallskip}\hline\noalign{\smallskip}
		MADDPG   & \textbf{4.879}($\pm$1.231)   & 6.192($\pm$2.213) &  618.641($\pm$32.796)\\
		\noalign{\smallskip}\hline\noalign{\smallskip}
		MAPPO   & 4.875($\pm$1.312)   & 6.819 ($\pm$2.131) &  619.89($\pm$24.139)\\
		\noalign{\smallskip}\hline\noalign{\smallskip}
		MARL-CAVG   &  4.265($\pm$1.913)  & \textbf{3.123}($\pm$1.139) & \textbf{669.119}($\pm$28.895)  \\
		\noalign{\smallskip}\hline\hline
	\end{tabular}
	\end{adjustbox}
	\label{eighttable}
\end{table}

\noindent \textbf{Evaluation in Merge Control}
 In the \textit{merge} scenario, the number of controlled and uncontrolled vehicles varies with time due to the inflow and outflow. MARL-CAVG method treats it through a limited multi-agent setting, transforming a tremendous state space using graph attention mechanism to handle the varying feature vector size. In this case, we can also find that our model outperforms the baselines in most evaluation indicators.

%  We find that MARL-CAVG method has the smallest deterioration when introducing more traffic inflow.

%Define merge: https://github.com/flow-project/flow/blob/a511c41c48e6b928bb2060de8ad1ef3c3e3d9554/flow/envs/multiagent/merge.py#L19

\begin{table}[H]
   
	\centering
	\caption{Performance comparison in merge network}
	\begin{adjustbox}{max width=0.45\textwidth}
	\begin{tabular}{cccccc}\hline\hline \noalign{\smallskip}
		Methods & Velocity ($m/s$) & Acc (0.1 $m^2/s$) & Return \\
		\noalign{\smallskip}\hline\noalign{\smallskip}
		IDM   & 2.881 & 6.386  &339.24\\
		\noalign{\smallskip}\hline\noalign{\smallskip}
		DDPG   &  4.512($\pm$1.021) &3.87 ($\pm$1.013)    &318.327($\pm$18.149)\\
		\noalign{\smallskip}\hline\noalign{\smallskip}
		PPO   & 4.663($\pm$2.183)  & 3.88($\pm$1.216)  & 321.415($\pm$15.291) \\
		\noalign{\smallskip}\hline\noalign{\smallskip}
		MADDPG   & 4.718($\pm$1.132)  & 4.123($\pm$1.516)    &325.659($\pm$12.261)\\
		\noalign{\smallskip}\hline\noalign{\smallskip}
		MAPPO   & \textbf{4.723($\pm$1.912)}  &3.298($\pm$1.158)    &365.732($\pm$12.261)\\
		\noalign{\smallskip}\hline\noalign{\smallskip}
		MARL-CAVG   & 4.719($\pm$1.294)   & \textbf{2.213($\pm$1.391)}  & 398.220($\pm$9.5703) \\
		\noalign{\smallskip}\hline\hline
	\end{tabular}
	\end{adjustbox}
	\label{neighbor}

\end{table}

\subsection{Further discussion}

\subsubsection{Visualization of Control Performance}
\label{visualizationcontrol}
To evaluate the control performance, we select the \textit{ring} network as an example and plot space-time diagram and velocity figures with the trained policy after 200 episodes. The number of head in attention module is 8. We test each method with 200 time steps and a target speed of 20km/h, then record the average speed for all the vehicles in the current road network.

\begin{figure}[htbp]
\centering
\includegraphics[width=8cm]{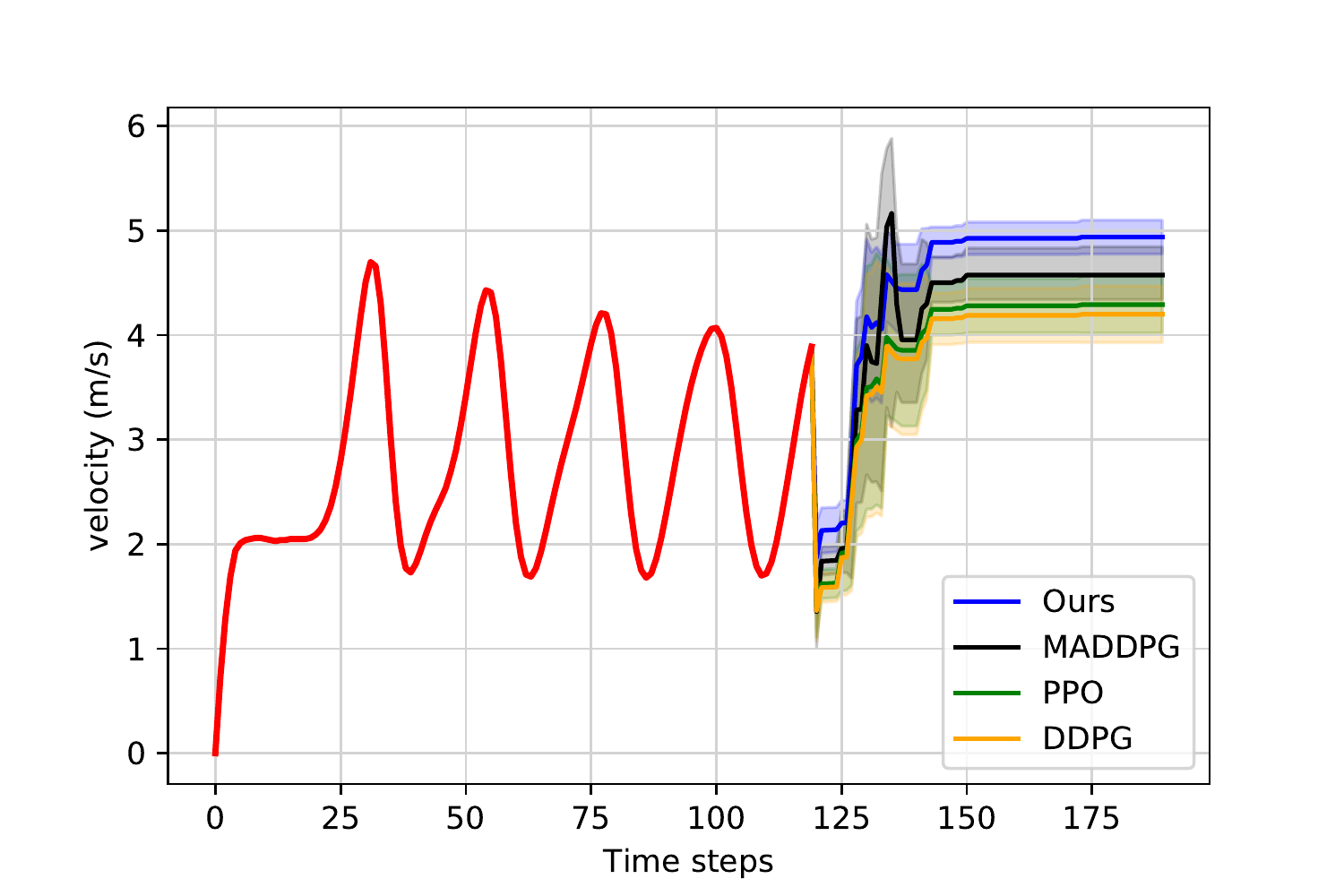}
\caption{Velocity performance in ring network. The red curve stands for the control performance with all human driving vehicles which is unstable. After automation turned on, the flow becomes stable.}
\label{ringv}
\end{figure}

From the result in Fig.~\ref{ringv}, we can see that after automation turns on, the velocity will become stable. Furthermore, our model can reach the highest speed compared to other baselines.

\begin{figure}[htbp]
\centering
\includegraphics[width=7cm]{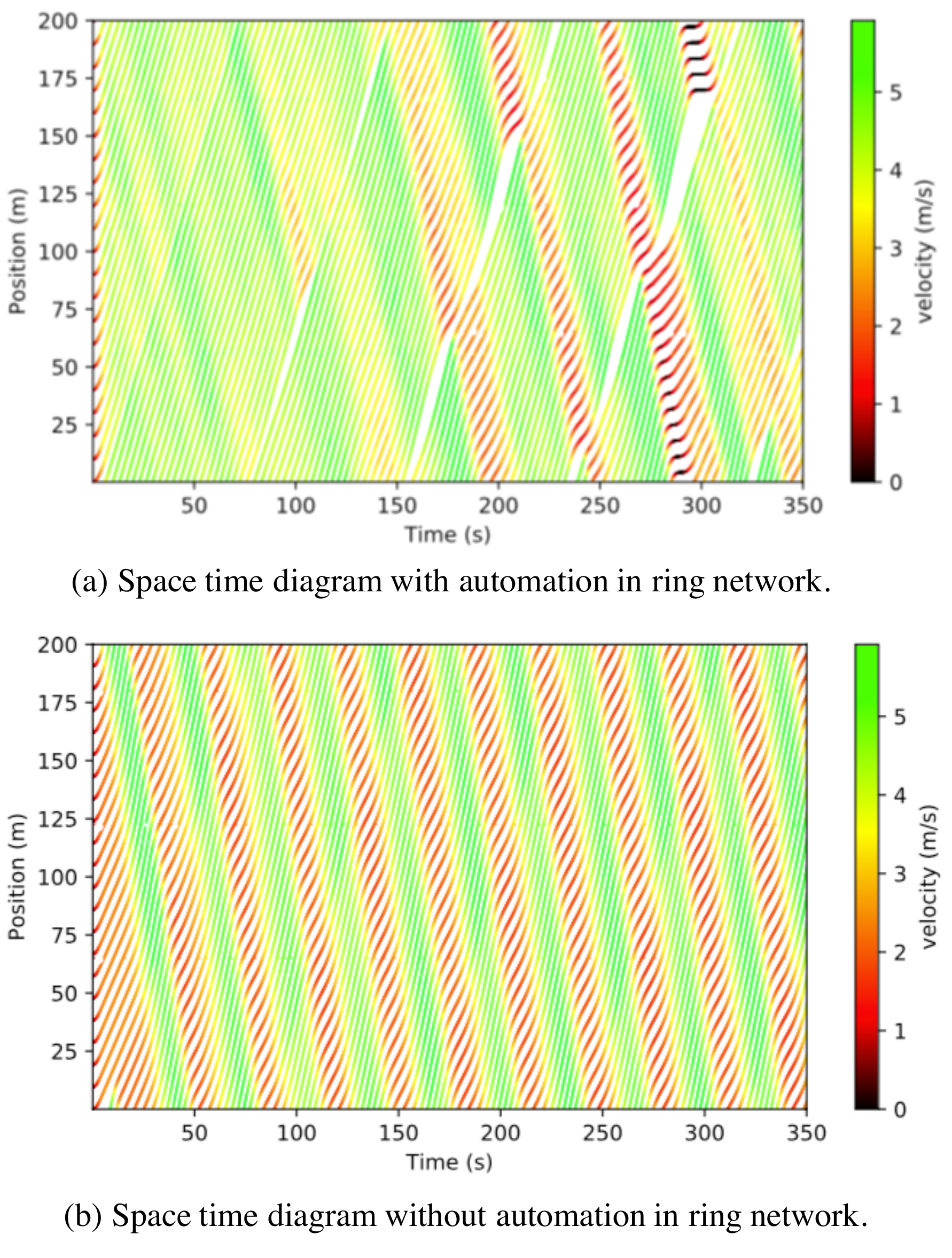}
\caption{Space time diagram with and without automation in the ring network. Different colors stand for different velocities. We can see that without automation, the velocity is not smooth, which is due to the stop and go effect, while with automation turned on, the automated vehicles can mitigate shock-wave.}
\label{st}
\end{figure}

To visualize the impact of shock-wave, we further compare the space-time diagram performance in the ring network before and after the automation is turned on. We can see from Fig.~\ref{st}-(b) that the velocity fluctuates sharply. With automation turned on, the velocity becomes smooth, and the average velocity increases, as shown in Fig.~\ref{st}-(a). 

These findings can demonstrate that while all the models can learn to regulate the traffic, the MARL-CAVG model can achieve the overall best performance.

% \begin{figure}[htbp]
% \centering
% \includegraphics[width=9cm]{picture/stdiagram-notitle.pdf}
% \caption{Space time diagram with automation in ring network. We can see that with automation turned, the autonomous can mitigate shock-wave.}
% \label{withauto}
% \end{figure}

% \begin{figure}[htbp]
% \centering
% \includegraphics[width=8cm]{picture/velocity8.pdf}
% \caption{Velocity performance in intersection network}
% \label{8v}
% \end{figure}

% \begin{table*}[htbp] \small
% \centering  
% \caption{Methods comparison within different scenarios.}
% \begin{tabular}{c c c c c c c c c c c } 
% \toprule[1.5pt]
% Method & Ring-v0 & Ring-v1 & Ring-v2 &   Eight-v0 & Eight-v1 & Eight-v2& Merge-v0 & Merge-v1 & Merge-v2 \\
% % \multicolumn{5}{c}{Transportation method}\\
% \midrule[1pt]
% IDM  & 424.12 & 424.12 & 424.12 & 37.34 &33.65 &40.29 &1157.46&2573.90&1157.46 \\
% % \multicolumn{5}{c}{$\sigma_s = 0.075$}\\
% \midrule[1pt]
% PPO  & -170 & 74.33 & 45.92 & 0.26  \\
% DDPG & -156 & 78.08 & 45.60 & 0.26 \\
% \midrule[1pt]
% MAPPO  & -170 & 74.33 & 45.92 & 0.26  \\
% MADDPG & -156 & 78.08 & 45.60 & 0.26 \\
% \midrule[1pt]
% Ours  & -170 & 74.33 & 45.92 & 0.26  \\
% \bottomrule[1.5pt]
% \end{tabular}
% \label{table3}
% \end{table*}

\subsubsection{Evaluation of different penetration rates}

\begin{figure}[H]
\centering
\includegraphics[width=9cm]{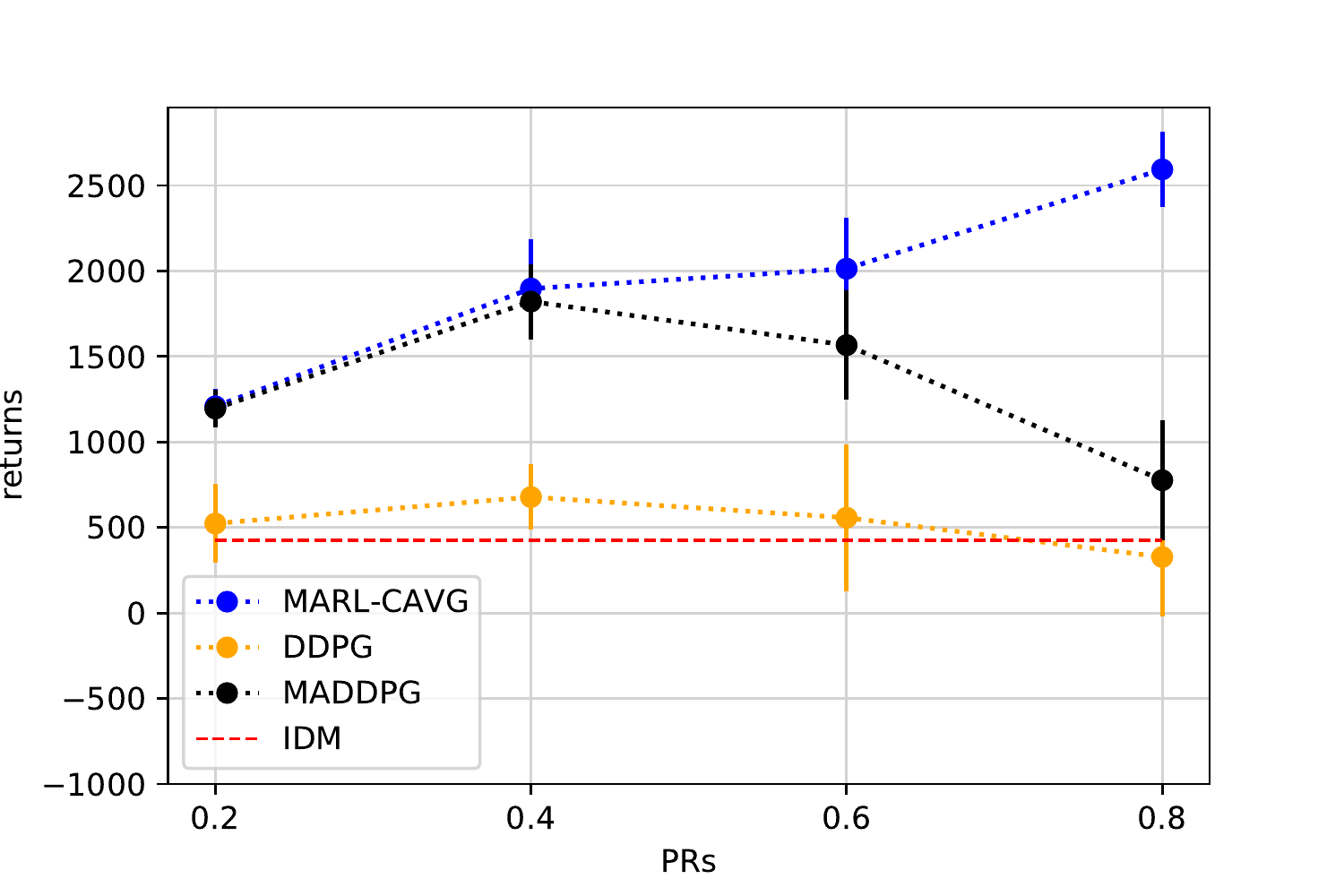}
\caption{Evaluation of returns under different penetration rates (PRs) in ring scenario. We fix the parameter of experiment in Section~\ref{visualizationcontrol} and only change different PRs.}
\label{pr}
\end{figure}

%%shared  parameter is easy to train however non-shared will make it harder
The penetration rate is a critical parameter that can affect the model's performance. To evaluate the performance of our model under different penetration rates, we select the typical ring scenario to make the comparison. We select the typical multi-agent RL approach (MADDPG) and single-agent RL approach (DDPG) as the baselines. As shown in Fig.~\ref{pr}, we can see that with the increase of penetration rates, the return of both the single-agent and multi-agent approaches first increases then decreases. The reason is that with more automated vehicles in the road network, there is a larger control policy space to explore, which hinders the training efficiency. Owing to the parameter sharing and graph attention within a certain scan scale, the MARL-CAVG can efficiently handle the increased number of controlled agents, and therefore achieve increasing return and the best overall performance.

% \begin{table}[htbp]
% 	\centering
% 	\footnotesize
% 	\caption{Performance comparison with different penetration rates in ring network}
% 	\begin{adjustbox}{max width=0.45\textwidth}

% 	\begin{tabular}{cccccc}\hline\hline \noalign{\smallskip}
% 		Methods  &PR &Velocity ($m/s$) &Acc (0.1 $m^2/s$) & Return\\ %
% 		\noalign{\smallskip}\hline\noalign{\smallskip}
% 		IDM   & &2.754   & 3.318 &424.12\\
% 		\noalign{\smallskip}\hline\noalign{\smallskip}
% 		DDPG-1  & &3.134 ($\pm$0.148)  & 2.718($\pm$0.378) &-70.063($\pm$70.506)  \\
% 		\noalign{\smallskip}\hline\noalign{\smallskip}
% 		DDPG-2  & &3.134 ($\pm$0.148)  & 2.718($\pm$0.378) &-70.063($\pm$70.506)  \\
%     	\noalign{\smallskip}\hline\noalign{\smallskip}
% 		PPO   & 3.165 ($\pm$0.145)  & 3.129 ($\pm$0.369)&-676.959($\pm$65.046) \\
% 		\noalign{\smallskip}\hline\noalign{\smallskip}
% 		MADDPG   & 3.270 ($\pm$0.148)   &2.121 ($\pm$0.366) &779.140($\pm$29.178)  \\
% 		\noalign{\smallskip}\hline\noalign{\smallskip}
% 		MAPPO   &  3.379 ($\pm$0.142) & 2.782 ($\pm$0.325) &776.225($\pm$20.121) \\
% 		\noalign{\smallskip}\hline\noalign{\smallskip}
% 		MARL-CAVG   & \textbf{3.391} ($\pm$0.092)  & \textbf{0.835 }($\pm$0.171) &\textbf{2593.99}($\pm$51.820)  \\
% 		\noalign{\smallskip}\hline\hline
% 	\end{tabular}
% 	\end{adjustbox}
% 	\label{ringpenetration}
% \end{table}

\subsubsection{Evaluation of different target speeds}
%%training on specific test on different target speeds
% 

\begin{figure}[htbp]
\centering
\includegraphics[width=9cm]{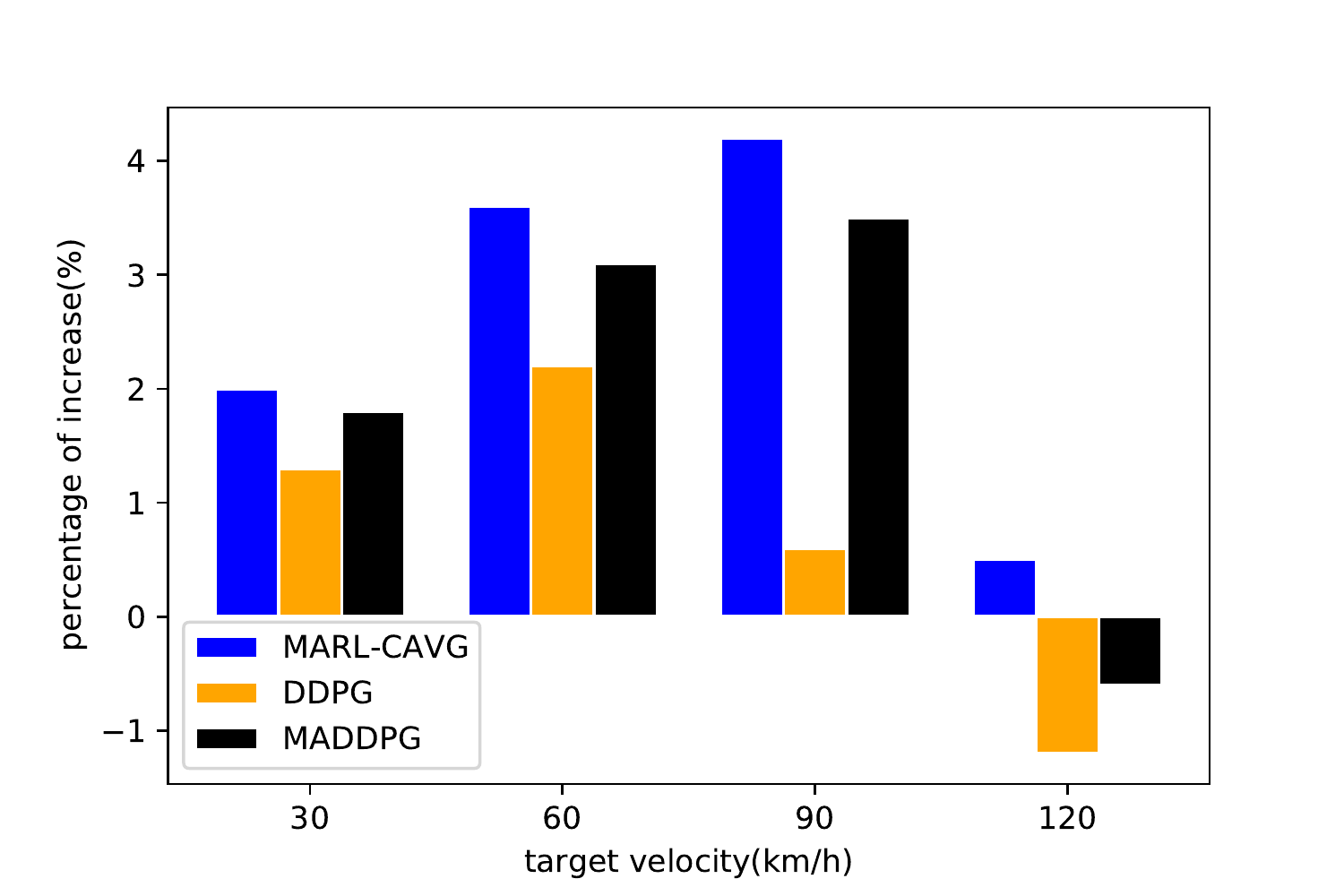}
\caption{Evaluation of percentage increase of returns under different target speeds in ring scenario. We fix the parameter of experiment in Section~\ref{visualizationcontrol} and only change different target speeds. We select the 20km/h as the baseline for each method and try with different target speeds.}
\label{ts}
\end{figure}

In real-world scenarios, different roads have different speed limits. Under some advanced freeway management frameworks, the speed limits can change over time \cite{wu2020differential}. In our study, we train the agents with a fixed target speed (20 km/h). It is curious to explore whether the agents can generalize to the scenarios with different speed limits (e.g., whether agents trained on urban roads generalizable to freeways).  In this study, we fix the penetration rate as 0.4, and try with different target speeds, then evaluate different methods' performance. We set the target speed 20km/h during training and calculate the percentage(\%) of reward increase for each method based on their 20km/h baseline. As shown in Fig.~\ref{ts}, we can find that for each method, the increase of target velocity leads to increase of collected reward. However, when the target velocity is too high (e.g., 120km/h), then the agent's performance will decrease. Our model achieves the best return given the highest target velocity. This demonstrates that our model is more generalizable to scenarios with different speed limits.

\subsubsection{Evaluation of different architectures}

\begin{figure}[H]
\centering
\includegraphics[width=9cm]{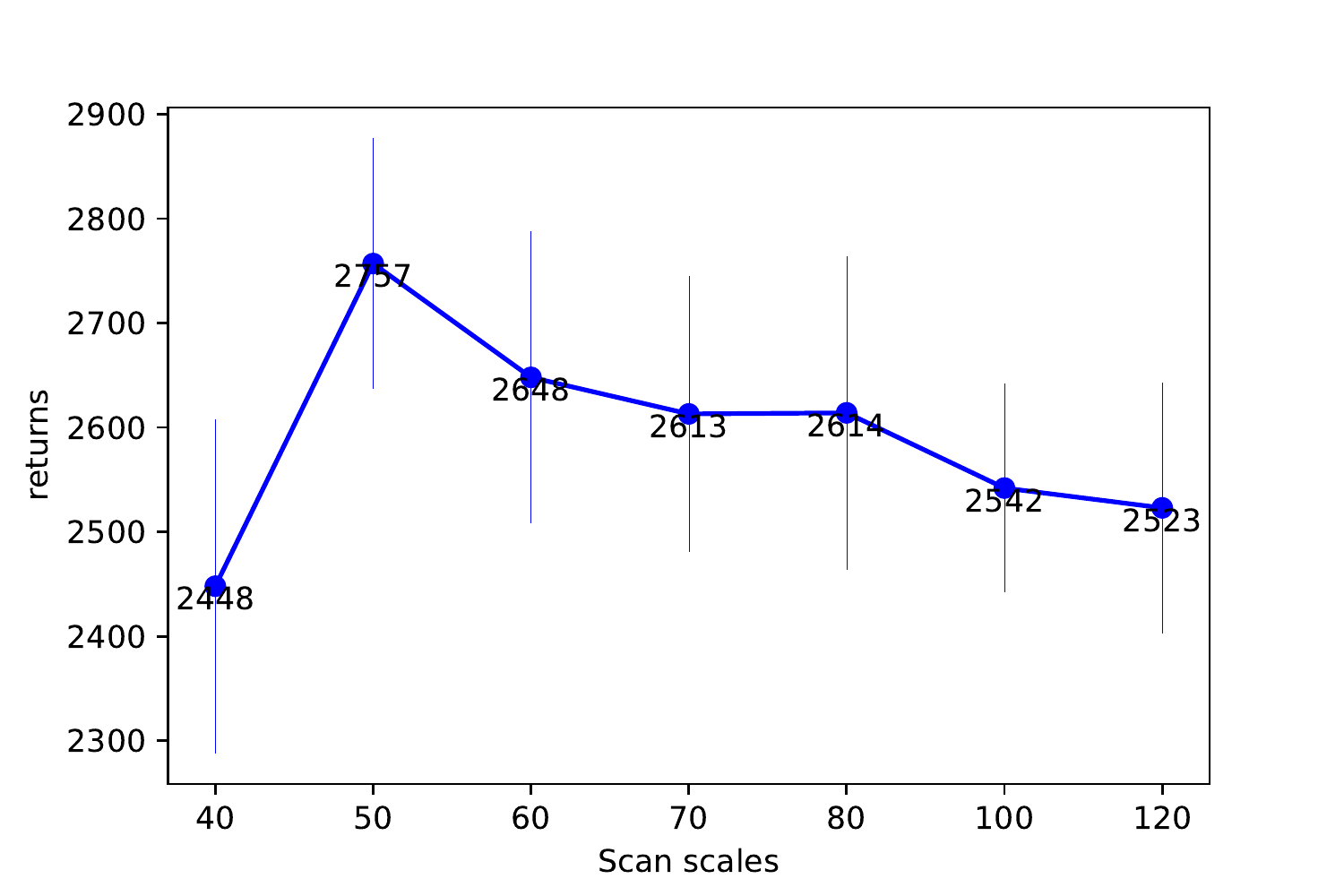}
\caption{Evaluation of returns under different scan scales(SCs) in ring scenario. We fix the parameter of experiment in Section~\ref{visualizationcontrol} and only change different SCs.}
\label{SC}
\end{figure}

For different architectures of the model, we first evaluate the effects of the range of scan scales on model performance. The results are shown in Figure~\ref{SC}. From the results, we can find that if we slightly enlarge the scan scale, the performance will be better because it can include more neighboring information. However, further increase of scan scale will decrease the model's performance because more redundant information will make learning becomes harder.

Furthermore, we evaluate the performance in terms of different information used to build the adjacency matrix. We consider only speed information, position information, and both speed and position information as given in Equation~\ref{feature}. The different information considered are as follows:

Only consider relative position information:
\begin{equation}
 m_i=x_i-x_j,
\label{pos}
\end{equation}
where $x_i$ and $x_j$ are the position  of the $i$-th agent and its surrounding $j$-th agent.

Only consider velocity information:
\begin{equation}
 m_i=\frac{v_T}{v_i(|v_j-v_i|)+\epsilon},
\label{vel}
\end{equation}
where $v_i$ and $v_j$ are the velocity of $i$-th agent and its surrounding $j$-th agent, $v_T$ is the target speed, and $\epsilon$ is a small positive number to keep it from divided by 0.

The philosophy of designing these information representations is, if two vehicles are running slowly on the road and are far away from each other, their correlation should be weak so that the measure should be larger. On the contrary, if a vehicle is running fast along with a slow vehicle and they are very closed, then they are more likely to be affected by each other, either because of the safety or efficiency consideration. Therefore, the measure will be smaller, and priority will be higher. %两个车离得近，就比较重视

\begin{table}[H]
	\centering
	\footnotesize
	\caption{Returns with different information in adjacency index}
	\begin{adjustbox}{max width=0.5\textwidth}

	\begin{tabular}{cccccc}\hline\hline \noalign{\smallskip}
		Adjacency index. & position & velocity & both \\ %
		\noalign{\smallskip}\hline\noalign{\smallskip}
		Returns   & 2490.99 ($\pm$ 20.149)   & 2601.87 ($\pm$ 19.825) &2710.32 ($\pm$ 23.581)  \\
		\noalign{\smallskip}\hline\hline
	\end{tabular}
	\end{adjustbox}

	\label{adj}
\end{table}

We fixed the penetration rate as 0.4, target speed as 30km/h. As shown in Table~\ref{adj},  speed information is more important than position information. Integrating both position and velocity information through velocity field can achieve the best overall performance. 

\subsubsection{Evaluation of attention module}

To evaluate the effectiveness of the attention setting, we conduct experiments with/without the attention module and a different number of heads in the attention module. In the experiment, the penetration rate is 0.4; the target speed is 30km/h. 

\begin{table}[H]
	\centering
	\footnotesize
	\caption{Returns with different heads in attention module}
	\begin{adjustbox}{max width=0.45\textwidth}

	\begin{tabular}{cccccc}\hline\hline \noalign{\smallskip}
		Heads & Returns \\ %
		\noalign{\smallskip}\hline\noalign{\smallskip}
		0  & 2423.19 ($\pm$39.193)   \\
		\noalign{\smallskip}\hline\noalign{\smallskip}
		2   & 2515.89 ($\pm$48.131)   \\
		\noalign{\smallskip}\hline\noalign{\smallskip}
		4   & 2566.23 ($\pm$43.123)  \\
		\noalign{\smallskip}\hline\noalign{\smallskip}
		6   & 2586.23 ($\pm$41.641)    \\
		\noalign{\smallskip}\hline\noalign{\smallskip}
		8   & 2624.20 ($\pm$41.213)    \\
		\noalign{\smallskip}\hline\noalign{\smallskip}
		10   &  2516.10 ($\pm$39.142) \\
		\noalign{\smallskip}\hline\hline
	\end{tabular}
	\end{adjustbox}
	\label{attention}
\end{table}
From Table~\ref{attention}, we can find that without attention module, the performance of the model decreases a lot. With the increase of attention heads, it will increase model's performance, while too many heads (heads$\geq$10) will decrease the performance.

\section{Conclusions and Discussion} \label{sec:conclusion}

In this paper, we propose a graph convolutional reinforcement learning approach for CAV control by encouraging cooperation for efficient traffic control in mixed autonomy.

We conduct extensive experiments based on different road networks and demonstrate the superior performance of our proposed MARL-CAVG method over both reinforcement learning and existing traffic flow simulation baselines. There are two major findings worth noting. First, we find that multi-agent training with the shared policy can achieve much better performance than those single-agent training strategies. Second, efficient communication strategies, such as the graph attention on surrounding neighbors proposed in this paper, can significantly enhance the cooperation among agents, which improves both efficiency and safety of the system. Overall, our method can achieve the best performance under different road networks, target speeds, penetration rates.  Our findings provide valuable insights into the design of the connected and automated driving system.

There are several directions for future research and improvements. In particular, as multi-agent training is quite unstable, a small change in the environment setting will result in a large return shift. Thus, it is critical to explore how to better stabilize the training in dynamic settings. In the future, we will also try to develop sim-to-real transfer learning \cite{jang2019simulation} for mixed-autonomy control and implement our approach in real mobile robot vehicles.

% or


\ifCLASSOPTIONcaptionsoff
  \newpage
\fi



%\bibliographystyle{IEEEtran}

\bibliographystyle{IEEEtranN}
\bibliography{ref}

% Generated by IEEEtranN.bst, version: 1.14 (2015/08/26)
\begin{thebibliography}{30}
\providecommand{\natexlab}[1]{#1}
\providecommand{\url}[1]{#1}
\csname url@samestyle\endcsname
\providecommand{\newblock}{\relax}
\providecommand{\bibinfo}[2]{#2}
\providecommand{\BIBentrySTDinterwordspacing}{\spaceskip=0pt\relax}
\providecommand{\BIBentryALTinterwordstretchfactor}{4}
\providecommand{\BIBentryALTinterwordspacing}{\spaceskip=\fontdimen2\font plus
\BIBentryALTinterwordstretchfactor\fontdimen3\font minus
  \fontdimen4\font\relax}
\providecommand{\BIBforeignlanguage}[2]{{%
\expandafter\ifx\csname l@#1\endcsname\relax
\typeout{** WARNING: IEEEtranN.bst: No hyphenation pattern has been}%
\typeout{** loaded for the language `#1'. Using the pattern for}%
\typeout{** the default language instead.}%
\else
\language=\csname l@#1\endcsname
\fi
#2}}
\providecommand{\BIBdecl}{\relax}
\BIBdecl

\bibitem[Treiber and Kesting(2013)]{treiber2013traffic}
M.~Treiber and A.~Kesting, ``Traffic flow dynamics,'' \emph{Traffic Flow
  Dynamics: Data, Models and Simulation, Springer-Verlag Berlin Heidelberg},
  2013.

\bibitem[Abbeel et~al.(2007)Abbeel, Coates, Quigley, and
  Ng]{abbeel2007application}
P.~Abbeel, A.~Coates, M.~Quigley, and A.~Y. Ng, ``An application of
  reinforcement learning to aerobatic helicopter flight,'' in \emph{Advances in
  neural information processing systems}, 2007, pp. 1--8.

\bibitem[Silver et~al.(2017)Silver, Schrittwieser, Simonyan, Antonoglou, Huang,
  Guez, Hubert, Baker, Lai, Bolton, et~al.]{silver2017mastering}
D.~Silver, J.~Schrittwieser, K.~Simonyan, I.~Antonoglou, A.~Huang, A.~Guez,
  T.~Hubert, L.~Baker, M.~Lai, A.~Bolton \emph{et~al.}, ``Mastering the game of
  go without human knowledge,'' \emph{nature}, vol. 550, no. 7676, pp.
  354--359, 2017.

\bibitem[Wang et~al.(2018)Wang, Chan, and
  de~La~Fortelle]{wang2018reinforcement}
P.~Wang, C.-Y. Chan, and A.~de~La~Fortelle, ``A reinforcement learning based
  approach for automated lane change maneuvers,'' in \emph{2018 IEEE
  Intelligent Vehicles Symposium (IV)}.\hskip 1em plus 0.5em minus 0.4em\relax
  IEEE, 2018, pp. 1379--1384.

\bibitem[Wu et~al.(2017)Wu, Kreidieh, Parvate, Vinitsky, and Bayen]{wu2017flow}
C.~Wu, A.~Kreidieh, K.~Parvate, E.~Vinitsky, and A.~M. Bayen, ``Flow:
  Architecture and benchmarking for reinforcement learning in traffic
  control,'' \emph{arXiv preprint arXiv:1710.05465}, 2017.

\bibitem[Rasekhipour et~al.(2016)Rasekhipour, Khajepour, Chen, and
  Litkouhi]{rasekhipour2016potential}
Y.~Rasekhipour, A.~Khajepour, S.-K. Chen, and B.~Litkouhi, ``A potential
  field-based model predictive path-planning controller for autonomous road
  vehicles,'' \emph{IEEE Transactions on Intelligent Transportation Systems},
  vol.~18, no.~5, pp. 1255--1267, 2016.

\bibitem[Luo et~al.(2019)Luo, Yang, Xu, Qin, and Li]{luo2019cooperative}
Y.~Luo, G.~Yang, M.~Xu, Z.~Qin, and K.~Li, ``Cooperative lane-change maneuver
  for multiple automated vehicles on a highway,'' \emph{Automotive Innovation},
  vol.~2, no.~3, pp. 157--168, 2019.

\bibitem[Lopez et~al.(2018)Lopez, Behrisch, Bieker-Walz, Erdmann,
  Fl{\"o}tter{\"o}d, Hilbrich, L{\"u}cken, Rummel, Wagner, and
  WieBner]{lopez2018microscopic}
P.~A. Lopez, M.~Behrisch, L.~Bieker-Walz, J.~Erdmann, Y.-P. Fl{\"o}tter{\"o}d,
  R.~Hilbrich, L.~L{\"u}cken, J.~Rummel, P.~Wagner, and E.~WieBner,
  ``Microscopic traffic simulation using sumo,'' in \emph{IEEE International
  Conference on Intelligent Transportation Systems (ITSC)}, 2018, pp.
  2575--2582.

\bibitem[Shi et~al.(2019)Shi, Wang, Cheng, Chan, and Huang]{shi2019driving}
T.~Shi, P.~Wang, X.~Cheng, C.-Y. Chan, and D.~Huang, ``Driving decision and
  control for autonomous lane change based on deep reinforcement learning,''
  \emph{arXiv preprint arXiv:1904.10171}, 2019.

\bibitem[Shalev-Shwartz et~al.(2016)Shalev-Shwartz, Shammah, and
  Shashua]{shalev2016safe}
S.~Shalev-Shwartz, S.~Shammah, and A.~Shashua, ``Safe, multi-agent,
  reinforcement learning for autonomous driving,'' \emph{arXiv preprint
  arXiv:1610.03295}, 2016.

\bibitem[Palanisamy(2019)]{palanisamy2019multi}
P.~Palanisamy, ``Multi-agent connected autonomous driving using deep
  reinforcement learning,'' \emph{arXiv preprint arXiv:1911.04175}, 2019.

\bibitem[Wang et~al.(2020)Wang, Hu, Li, and Li]{wang2019cooperative}
G.~Wang, J.~Hu, Z.~Li, and L.~Li, ``Harmonious lane changing via deep
  reinforcement learning,'' \emph{IEEE Transactions on Intelligent
  Transportation Systems}, no. accepted, 2020.

\bibitem[Wu et~al.(2020{\natexlab{a}})Wu, Pan, Chen, Long, Zhang, and
  Philip]{wu2020comprehensive}
Z.~Wu, S.~Pan, F.~Chen, G.~Long, C.~Zhang, and S.~Y. Philip, ``A comprehensive
  survey on graph neural networks,'' \emph{IEEE Transactions on Neural Networks
  and Learning Systems}, 2020.

\bibitem[Iqbal and Sha(2019)]{iqbal2019actor}
S.~Iqbal and F.~Sha, ``Actor-attention-critic for multi-agent reinforcement
  learning,'' in \emph{International Conference on Machine Learning}, 2019, pp.
  2961--2970.

\bibitem[Agarwal et~al.(2019)Agarwal, Kumar, and Sycara]{agarwal2019learning}
A.~Agarwal, S.~Kumar, and K.~Sycara, ``Learning transferable cooperative
  behavior in multi-agent teams,'' \emph{arXiv preprint arXiv:1906.01202},
  2019.

\bibitem[Jiang et~al.(2018)Jiang, Dun, and Lu]{jiang2018graph}
J.~Jiang, C.~Dun, and Z.~Lu, ``Graph convolutional reinforcement learning for
  multi-agent cooperation,'' \emph{arXiv preprint arXiv:1810.09202}, vol.~2,
  no.~3, 2018.

\bibitem[Kreidieh et~al.(2018)Kreidieh, Wu, and Bayen]{kreidieh2018dissipating}
A.~R. Kreidieh, C.~Wu, and A.~M. Bayen, ``Dissipating stop-and-go waves in
  closed and open networks via deep reinforcement learning,'' in \emph{2018
  21st International Conference on Intelligent Transportation Systems
  (ITSC)}.\hskip 1em plus 0.5em minus 0.4em\relax IEEE, 2018, pp. 1475--1480.

\bibitem[Treiber et~al.(2000)Treiber, Hennecke, and
  Helbing]{treiber2000congested}
M.~Treiber, A.~Hennecke, and D.~Helbing, ``Congested traffic states in
  empirical observations and microscopic simulations,'' \emph{Physical review
  E}, vol.~62, no.~2, p. 1805, 2000.

\bibitem[Sutton and Barto(2018)]{sutton2018reinforcement}
R.~S. Sutton and A.~G. Barto, \emph{Reinforcement learning: An
  introduction}.\hskip 1em plus 0.5em minus 0.4em\relax MIT press, 2018.

\bibitem[Wei et~al.(2019)Wei, Xu, Zhang, Zheng, Zang, Chen, Zhang, Zhu, Xu, and
  Li]{wei2019colight}
H.~Wei, N.~Xu, H.~Zhang, G.~Zheng, X.~Zang, C.~Chen, W.~Zhang, Y.~Zhu, K.~Xu,
  and Z.~Li, ``Colight: Learning network-level cooperation for traffic signal
  control,'' in \emph{Proceedings of the 28th ACM International Conference on
  Information and Knowledge Management}, 2019, pp. 1913--1922.

\bibitem[Zhang et~al.(2020)Zhang, Zhu, Wang, and Xi]{zhang2020spatiotemporal}
C.~Zhang, J.~Zhu, W.~Wang, and J.~Xi, ``Spatiotemporal learning of multivehicle
  interaction patterns in lane-change scenarios,'' \emph{arXiv preprint
  arXiv:2003.00759}, 2020.

\bibitem[Vaswani et~al.(2017)Vaswani, Shazeer, Parmar, Uszkoreit, Jones, Gomez,
  Kaiser, and Polosukhin]{vaswani2017attention}
A.~Vaswani, N.~Shazeer, N.~Parmar, J.~Uszkoreit, L.~Jones, A.~N. Gomez,
  {\L}.~Kaiser, and I.~Polosukhin, ``Attention is all you need,'' in
  \emph{Advances in neural information processing systems}, 2017, pp.
  5998--6008.

\bibitem[Schulman et~al.(2017)Schulman, Wolski, Dhariwal, Radford, and
  Klimov]{schulman2017proximal}
J.~Schulman, F.~Wolski, P.~Dhariwal, A.~Radford, and O.~Klimov, ``Proximal
  policy optimization algorithms,'' \emph{arXiv preprint arXiv:1707.06347},
  2017.

\bibitem[Vinitsky et~al.(2018)Vinitsky, Kreidieh, Le~Flem, Kheterpal, Jang, Wu,
  Wu, Liaw, Liang, and Bayen]{vinitsky2018benchmarks}
E.~Vinitsky, A.~Kreidieh, L.~Le~Flem, N.~Kheterpal, K.~Jang, C.~Wu, F.~Wu,
  R.~Liaw, E.~Liang, and A.~M. Bayen, ``Benchmarks for reinforcement learning
  in mixed-autonomy traffic,'' in \emph{Conference on Robot Learning}.\hskip
  1em plus 0.5em minus 0.4em\relax PMLR, 2018, pp. 399--409.

\bibitem[Sugiyama et~al.(2008)Sugiyama, Fukui, Kikuchi, Hasebe, Nakayama,
  Nishinari, Tadaki, and Yukawa]{sugiyama2008traffic}
Y.~Sugiyama, M.~Fukui, M.~Kikuchi, K.~Hasebe, A.~Nakayama, K.~Nishinari, S.-i.
  Tadaki, and S.~Yukawa, ``Traffic jams without bottlenecks—experimental
  evidence for the physical mechanism of the formation of a jam,'' \emph{New
  Journal of Physics}, vol.~10, no.~3, p. 033001, 2008.

\bibitem[Lillicrap et~al.(2015)Lillicrap, Hunt, Pritzel, Heess, Erez, Tassa,
  Silver, and Wierstra]{lillicrap2015continuous}
T.~P. Lillicrap, J.~J. Hunt, A.~Pritzel, N.~Heess, T.~Erez, Y.~Tassa,
  D.~Silver, and D.~Wierstra, ``Continuous control with deep reinforcement
  learning,'' \emph{arXiv preprint arXiv:1509.02971}, 2015.

\bibitem[Huang et~al.(2019)Huang, Braghin, and Arrigoni]{huang2019autonomous}
W.~Huang, F.~Braghin, and S.~Arrigoni, ``Autonomous vehicle driving via deep
  deterministic policy gradient,'' in \emph{ASME 2019 International Design
  Engineering Technical Conferences and Computers and Information in
  Engineering Conference}.\hskip 1em plus 0.5em minus 0.4em\relax American
  Society of Mechanical Engineers Digital Collection, 2019.

\bibitem[Lowe et~al.(2017)Lowe, Wu, Tamar, Harb, Abbeel, and
  Mordatch]{lowe2017multi}
R.~Lowe, Y.~I. Wu, A.~Tamar, J.~Harb, O.~P. Abbeel, and I.~Mordatch,
  ``Multi-agent actor-critic for mixed cooperative-competitive environments,''
  in \emph{Advances in neural information processing systems}, 2017, pp.
  6379--6390.

\bibitem[Wu et~al.(2020{\natexlab{b}})Wu, Tan, Qin, and
  Ran]{wu2020differential}
Y.~Wu, H.~Tan, L.~Qin, and B.~Ran, ``Differential variable speed limits control
  for freeway recurrent bottlenecks via deep actor-critic algorithm,''
  \emph{Transportation research part C: emerging technologies}, vol. 117, p.
  102649, 2020.

\bibitem[Jang et~al.(2019)Jang, Vinitsky, Chalaki, Remer, Beaver, Malikopoulos,
  and Bayen]{jang2019simulation}
K.~Jang, E.~Vinitsky, B.~Chalaki, B.~Remer, L.~Beaver, A.~A. Malikopoulos, and
  A.~Bayen, ``Simulation to scaled city: zero-shot policy transfer for traffic
  control via autonomous vehicles,'' in \emph{Proceedings of the 10th ACM/IEEE
  International Conference on Cyber-Physical Systems}, 2019, pp. 291--300.

\end{thebibliography}


\begin{IEEEbiography}[{\includegraphics[width=1in,height=1.25in,clip,keepaspectratio]{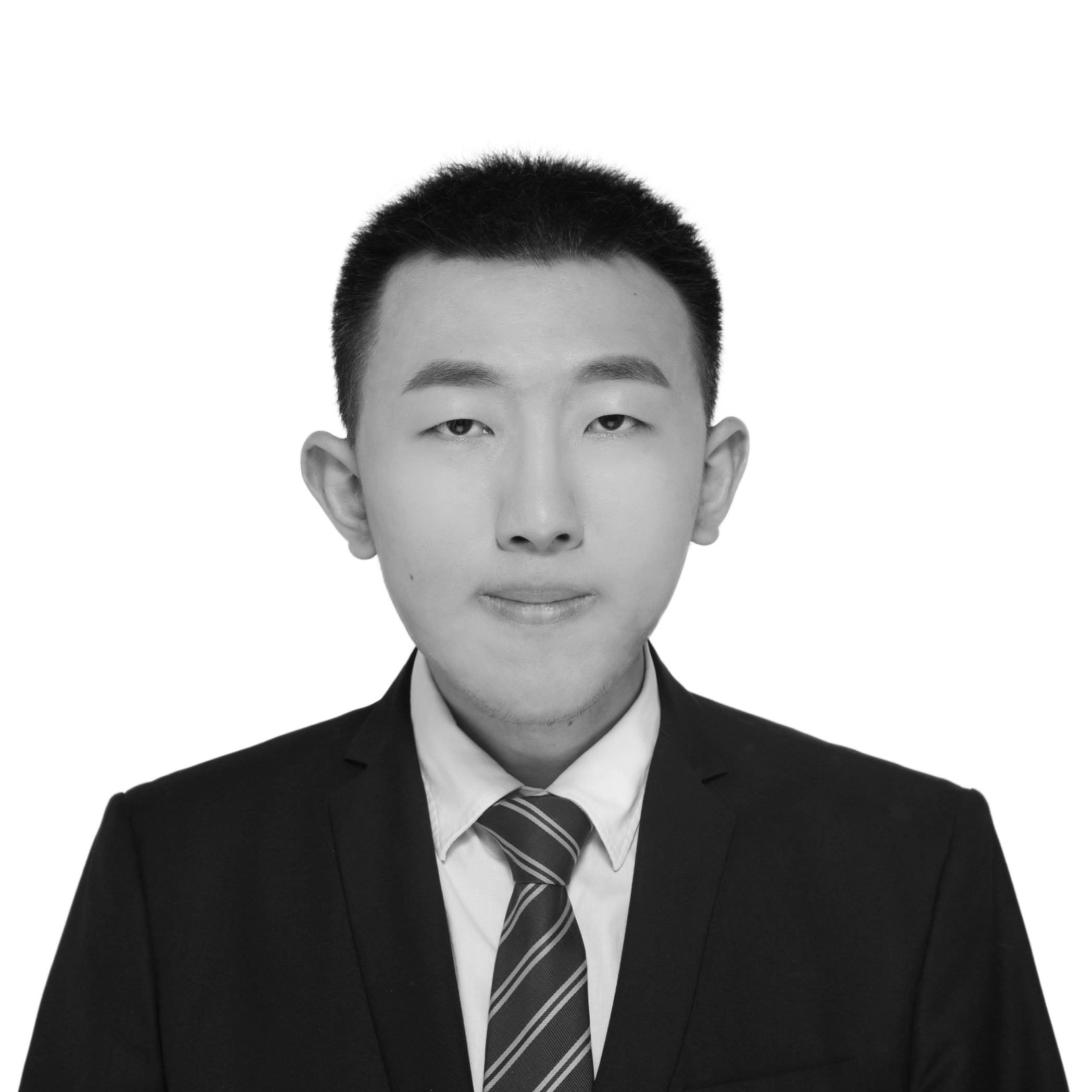}}]
{Tianyu Shi} received the B.S. degree in Vehicle Engineering from Beijing Institute of Technology Beijing, China, in 2019. He was a research assistant with Berkeley Deep Drive, University of California Berkeley.  He is now pursing master degree supervised by Lijun Sun, with Department of Civil Engineering at McGill University, supported by IVADO excellence scholarship program.
His current research centers on reinforcement learning, robotics and intelligent transportation systems.
\end{IEEEbiography}

\begin{IEEEbiography}[{\includegraphics[width=1in,height=1.25in,clip,keepaspectratio]{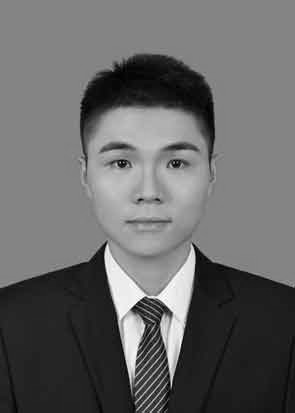}}]
{Jiawei Wang} received the B.S. degree in Traffic Engineering from Sun Yat-Sen University, Guangzhou, China, in 2016, and M.S. degree in Traffic information and control from Sun Yat-Sen University, Guangzhou, China, in 2019. He is now PhD candidate supervised by Lijun Sun, with Department of Civil Engineering at McGill University.
His current research centers on Intelligent transportation systems, traffic control and machine learning.
\end{IEEEbiography}

\begin{IEEEbiography}[{\includegraphics[width=1in,height=1.25in,clip,keepaspectratio]{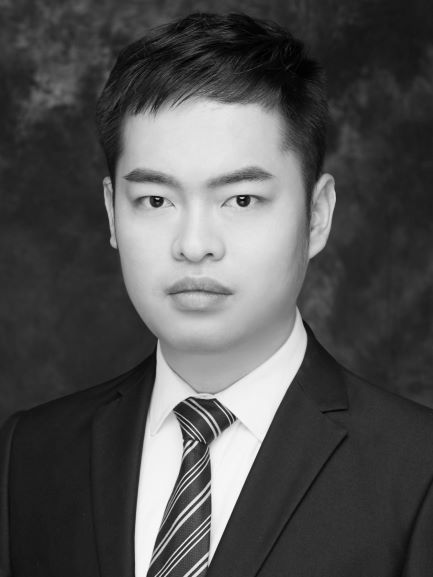}}]
{Yuankai Wu} received the PhD's degree from the School of Mechanical Engineering, Beijing Institute of Technology, Beijing, China, in 2019. He was a visit PhD student with Department of Civil \&
Environmental Engineering, University of Wisconsin-Madison from Nov. 2016 to Nov, 2017. He is a Postdoc researcher with Department of Civil Engineering at McGill University, supported by the Institute For Data Valorization (IVADO). His research interests include intelligent transportation systems, intelligent energy management and machine learning.
\end{IEEEbiography}

\begin{IEEEbiography}[{\includegraphics[width=1in,height=1.25in,clip,keepaspectratio]{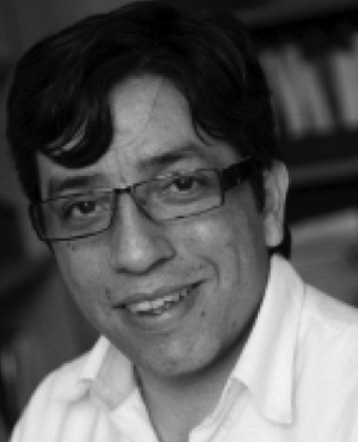}}]
{Luis Miranda-Moreno} received the Ph.D. degree in transportation engineering from the University of Waterloo, Waterloo, ON, Canada. He is currently an Associate Professor with the Department of Civil Engineering, McGill University. He is a specialist in transportation engineering, in the area of intelligent transportation systems, road safety, and the impact of transportation on the environment. He has extensive experience in the integration of emerging technologies and road safety methods for network screening, emerging surrogate safety analysis, and non-motorized transportation. He has worked on various projects for major cities, industries, and government agencies across North America. He has authored over 80 peer-reviewed journal articles. 
\end{IEEEbiography}

% \vspace{-90 mm}
\begin{IEEEbiography}[{\includegraphics[width=1in,height=1.25in,clip,keepaspectratio]{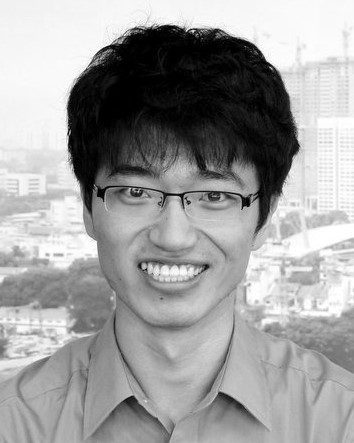}}]
{Lijun Sun} received the B.S. degree in Civil Engineering from Tsinghua University, Beijing, China, in 2011, and Ph.D. degree in Civil Engineering (Transportation) from National University of Singapore in 2015. He is currently an Assistant Professor with the Department of Civil Engineering at McGill University, Montreal, Quebec, Canada. His research centers on intelligent transportation systems, machine learning, spatiotemporal modeling, travel behavior, and agent-based simulation.
\end{IEEEbiography}


%\vfill


\end{document}